\documentclass{article}

\usepackage{microtype}
\usepackage{graphicx}
\usepackage{caption}
\usepackage{subcaption}
\usepackage{booktabs}

\usepackage{hyperref}

\usepackage[preprint]{icml2026}

\usepackage{times}
\usepackage{xspace}
\usepackage{url}
\usepackage{array}
\usepackage{tabularx}
\usepackage{multirow}
\usepackage{adjustbox}
\usepackage{listings}
\usepackage{listings}
\usepackage[table]{xcolor}
\usepackage[most]{tcolorbox}
\usepackage{enumitem}
\usepackage{float}

\usepackage{amsmath}
\usepackage{amssymb}
\usepackage{mathtools}
\usepackage{amsthm}

\usepackage{amsmath,amsfonts,bm}

\def\eqref#1{equation~\ref{#1}}

\def\1{\bm{1}}

\DeclareMathAlphabet{\mathsfit}{\encodingdefault}{\sfdefault}{m}{sl}
\SetMathAlphabet{\mathsfit}{bold}{\encodingdefault}{\sfdefault}{bx}{n}

\definecolor{mine_font}{RGB}{0, 128, 0}
\definecolor{lightblue}{RGB}{221, 232, 250}
\definecolor{tea_green}{RGB}{214, 234, 193}
\definecolor{hint_green}{RGB}{226,246,209}
\definecolor{Madang}{RGB}{190,235,159}
\definecolor{yellow_green}{RGB}{198,222,119}
\definecolor{link_water}{RGB}{221, 232, 250}
\definecolor{celestial_blue}{RGB}{52, 152, 219}
\definecolor{shakespeare}{RGB}{85, 154, 193}
\definecolor{buttermilk}{RGB}{255,242,174}
\definecolor{chardonnay}{RGB}{250,196,114}
\definecolor{rajah}{RGB}{253,180,98}
\definecolor{fog}{RGB}{213, 193, 234}
\definecolor{melon}{RGB}{254,191,181}
\definecolor{sundown}{RGB}{249, 180, 181}
\definecolor{mona_lisa}{RGB}{246,152,134}
\definecolor{salmon}{RGB}{242,131,107}
\definecolor{blue_x}{RGB}{142, 207, 201}
\definecolor{orange_x}{RGB}{255, 190, 122}

\definecolor{saltpan}{RGB}{238, 243, 232}
\definecolor{aqua_spring}{RGB}{232, 243, 232}
\definecolor{tea_green}{RGB}{214, 234, 193}
\definecolor{Madang}{RGB}{190,235,159}
\definecolor{fringy_flower}{RGB}{194, 234, 193}
\definecolor{aero_blue}{RGB}{193, 234, 213}
\definecolor{pixie_green}{RGB}{183,214,170}
\definecolor{french_pass}{RGB}{195,232,246}
\definecolor{ice_cold}{RGB}{169,232,220}
\definecolor{pale_turquoise}{RGB}{172,240,242}
\definecolor{cruise}{RGB}{179,226,205}
\definecolor{sail}{RGB}{163,205,235}
\definecolor{spindle}{RGB}{179,205,227}
\definecolor{link_water}{RGB}{221, 232, 250}
\definecolor{periwinkle}{RGB}{203,213,232}
\definecolor{zanah}{RGB}{220, 233, 213}
\definecolor{frostee}{RGB}{217, 231, 214}
\definecolor{opal}{RGB}{199, 221, 211}
\definecolor{jet_stream}{RGB}{188, 214, 210}
\definecolor{skeptic}{RGB}{153, 187, 167}
\definecolor{hint_green}{RGB}{226,246,209}
\definecolor{snow_flurry}{RGB}{230,245,201}
\definecolor{surf_crest}{RGB}{205,230,208}
\definecolor{yellow_green}{RGB}{198,222,119}
\definecolor{cream}{RGB}{255,255,204}
\definecolor{pale_prim}{RGB}{255,255,179}
\definecolor{spring_sun}{RGB}{242,243,195}
\definecolor{portafino}{RGB}{245,237,160}
\definecolor{buttermilk}{RGB}{255,242,174}
\definecolor{cream_brulee}{RGB}{255, 229, 151}
\definecolor{dairy_cream}{RGB}{254,226,189}
\definecolor{champagne}{RGB}{254,217,166}
\definecolor{chardonnay}{RGB}{250,196,114}
\definecolor{manhattan}{RGB}{226,180,125}
\definecolor{rajah}{RGB}{253,180,98}
\definecolor{early_dawn}{RGB}{252,243,218}
\definecolor{egg_shell}{RGB}{238, 234, 215}
\definecolor{selago}{RGB}{243, 232, 243}
\definecolor{quartz}{RGB}{219,223,238}
\definecolor{fog}{RGB}{213, 193, 234}
\definecolor{languid_lavender}{RGB}{222,203,228}
\definecolor{watusi}{RGB}{254,221,207}
\definecolor{coral_andy}{RGB}{243,204,205}
\definecolor{cosmos}{RGB}{248,209,210}
\definecolor{melon}{RGB}{254,191,181}
\definecolor{azalea}{RGB}{234, 193, 194}
\definecolor{beauty_bush}{RGB}{235, 185, 179}
\definecolor{sundown}{RGB}{249, 180, 181}
\definecolor{mona_lisa}{RGB}{246,152,134}
\definecolor{salmon}{RGB}{242,131,107}

\definecolor{summer_sky}{RGB}{58, 151, 233}
\definecolor{chateau_green}{RGB}{72, 179, 96}
\definecolor{matisse}{RGB}{25, 104, 167}
\definecolor{allports}{RGB}{31, 106, 125}
\definecolor{sun_shade}{RGB}{255, 144, 68}
\definecolor{flamingo}{RGB}{237, 88, 85}
\definecolor{studio}{RGB}{128, 91, 160}

\definecolor{maya_blue}{RGB}{102, 204, 255}
\definecolor{feijoa}{RGB}{178,223,138}
\definecolor{sushi}{RGB}{117, 168, 47}
\definecolor{norway}{RGB}{158, 194, 132}
\definecolor{japanese_laurel}{RGB}{53, 116, 40}
\definecolor{see_green}{RGB}{161,228,195}
\definecolor{monte_carlo}{RGB}{135,204,194}
\definecolor{granny_smith_apple}{RGB}{150,214,150}
\definecolor{moss_green}{RGB}{170,216,176}
\definecolor{chateau_green}{RGB}{72, 179, 96}
\definecolor{opal}{RGB}{164,207,190}
\definecolor{acapulco}{RGB}{117, 170, 148}
\definecolor{viridian}{RGB}{55, 137, 122}
\definecolor{amazon}{RGB}{56, 123, 84}
\definecolor{asparagus}{RGB}{123, 160, 91}
\definecolor{fruit_salad}{RGB}{91, 160, 94}
\definecolor{puerto_rico}{RGB}{72, 179, 150}
\definecolor{mountain_meadow}{RGB}{0, 163, 136}
\definecolor{matisse}{RGB}{25, 104, 167}
\definecolor{allports}{RGB}{31, 106, 125}
\definecolor{astral}{RGB}{55, 111, 137}
\definecolor{spring_leaves}{RGB}{46, 83, 117}
\definecolor{biscay}{RGB}{44, 62, 80}
\definecolor{midnight}{RGB}{0, 29, 50}
\definecolor{amethyst}{RGB}{153, 102, 204}
\definecolor{studio}{RGB}{128, 91, 160}
\definecolor{tapestry}{RGB}{194, 109, 132}
\definecolor{atomic_tangerine}{RGB}{255, 153, 102}
\definecolor{amber}{RGB}{255, 191, 0}
\definecolor{casablanca}{RGB}{244, 178, 84}
\definecolor{california}{RGB}{233, 140, 58}
\definecolor{tomato}{RGB}{255, 97, 56} 
\definecolor{alizarin}{RGB}{233, 58, 64}

\definecolor{linen}{RGB}{251, 239, 227}
\definecolor{double_pearl_lusta}{RGB}{253, 242, 208}
\definecolor{oasis}{RGB}{253, 242, 208}
\definecolor{milan}{RGB}{255, 254, 169}
\definecolor{texas}{RGB}{245, 232, 123}
\definecolor{maize}{RGB}{249, 212, 156}

\definecolor{turmeric}{RGB}{211, 178, 76}
\definecolor{saffron}{RGB}{249,193,62}
\definecolor{my_sin}{RGB}{255, 176, 59}
\definecolor{tree_poppy}{RGB}{246, 154, 27}
\definecolor{jaffa}{RGB}{240, 131, 58}
\definecolor{crusta}{RGB}{254, 127, 44}
\definecolor{tahiti_gold}{RGB}{223, 102, 36}
\definecolor{outrageous_orange}{RGB}{255, 100, 45}
\definecolor{safety_orange}{RGB}{254, 106, 0}

\definecolor{azalea}{RGB}{251, 196, 196}
\definecolor{oyster_pink}{RGB}{238,206,205} 
\definecolor{coral_candy}{RGB}{242,208,205} 
\definecolor{baby_pink}{RGB}{246, 194, 192}
\definecolor{petite_orchid}{RGB}{223, 157, 155}
\definecolor{apricot}{RGB}{241,140,122}
\definecolor{NY_pink}{RGB}{228,136,113}
\definecolor{carmine_pink}{RGB}{231, 76, 60}
\definecolor{deep_carmine_pink}{RGB}{236, 50, 67}

\definecolor{wewak}{RGB}{244, 143, 150}
\definecolor{light_coral}{RGB}{244, 127, 123}
\definecolor{bittersweet}{RGB}{255,111,105}
\definecolor{carnation}{RGB}{245, 80, 86}
\definecolor{flamingo}{RGB}{237, 88, 85}
\definecolor{sunset_orange}{RGB}{242,89,75}
\definecolor{ku_crimson}{RGB}{243, 0, 25}
\definecolor{amaranth}{RGB}{234,46,73}
\definecolor{valencia}{RGB}{214, 87, 70}
\definecolor{chilean_fire}{RGB}{215, 87, 44}
\definecolor{mexican_red}{RGB}{170, 41, 37}

\definecolor{napa}{RGB}{163, 154, 137}

\definecolor{athens_gray}{RGB}{236, 240, 241}
\definecolor{gallery}{RGB}{240,240,240}
\definecolor{mercury}{RGB}{230,230,230}
\definecolor{platinum}{RGB}{228,228,228}
\definecolor{silver}{RGB}{191,191,191}
\definecolor{aluminum}{RGB}{153,153,153}
\definecolor{ship_gray}{RGB}{77,77,77}
\definecolor{tuatara}{RGB}{67, 67, 67}

\definecolor{malibu}{RGB}{110, 180, 240}
\definecolor{celestial_blue}{RGB}{52, 152, 219}
\definecolor{curious_blue}{RGB}{41, 128, 185}
\definecolor{french_blue}{RGB}{0, 112, 182}
\definecolor{matisse}{RGB}{25, 104, 167}
\definecolor{shakespeare}{RGB}{85, 154, 193}
\definecolor{seagull}{RGB}{128,177,211}
\definecolor{jelly_bean}{RGB}{45, 126, 150}
\definecolor{venice_blue}{RGB}{87, 135, 105}
\definecolor{boston_blue}{RGB}{68, 147, 161}

\definecolor{turquoise}{RGB}{41,217,194}
\definecolor{java}{RGB}{2,190,196}
\definecolor{riptide}{RGB}{141,211,199}
\definecolor{mountain_meadow}{RGB}{0, 163, 136}
\definecolor{free_speech_aquamarine}{RGB}{0, 156, 114}

\definecolor{cosmic_latte}{RGB}{222, 247, 229}
\definecolor{chinook}{RGB}{163, 232, 178}
\definecolor{padua}{RGB}{121, 189, 143}
\definecolor{ocean_green}{RGB}{79, 176, 112}
\definecolor{pastel_green}{RGB}{107, 227, 135}
\definecolor{chateau_green}{RGB}{69, 191, 85}
\definecolor{RoyalBlue}{RGB}{69, 191, 85}
\definecolor{pigment_green}{RGB}{0, 175, 79}
\definecolor{fern}{RGB}{101,197,117}
\definecolor{killarney}{RGB}{56, 113, 66}

\newcommand{\numrwe}[2]{%
    \begin{tikzpicture}[baseline]
        \pgfmathparse{#1 > #2 ? 1 : 0}
        \ifnum\pgfmathresult=0
            \pgfmathsetmacro{\percentdiff}{min(130, 130*(#2-#1)/#2)}
            \pgfmathsetmacro{\intensity}{\percentdiff}
            \fill[monte_carlo!\intensity!white, rounded corners=1] (-0.6em, -0.3em) rectangle (2.6em, 1em);
        \fi
        \node[inner sep=0pt] at (1em, 0.7ex) {#1};
    \end{tikzpicture}%
}

\newcolumntype{L}{>{\hspace{1.5pt}}l<{\hspace{1.5pt}}}
\newcolumntype{C}{>{\hspace{1.5pt}}c<{\hspace{1.5pt}}}
\newcolumntype{Y}{>{\hsize=2.2\hsize\raggedright\arraybackslash}X}
\newcolumntype{Z}{>{\hsize=0.85\hsize\centering\arraybackslash}X}

\makeatletter
\renewcommand{\printAffiliationsAndNotice}[1]{\global\icml@noticeprintedtrue%
  \stepcounter{@affiliationcounter}%
  {\let\thefootnote\relax\footnotetext{\hspace*{-\footnotesep}\ificmlshowauthors #1\fi%
      \forloop{@affilnum}{1}{\value{@affilnum} < \value{@affiliationcounter}}{
        \textsuperscript{\arabic{@affilnum}}\ifcsname @affilname\the@affilnum\endcsname%
          \csname @affilname\the@affilnum\endcsname%
        \else
          {\bf AUTHORERR: Missing \textbackslash{}icmlaffiliation.}
        \fi
      }. \textsuperscript{\dag}Corresponding authors.

      \ \\
      \Notice@String
    }
  }
}
\makeatother

\icmltitlerunning{EtCon: Edit then Consolidate for Reliable Knowledge Editing}

\begin{document}

\twocolumn[
  \icmltitle{EtCon: Edit then Consolidate for Reliable Knowledge Editing}

  \icmlsetsymbol{cor}{\dag}

  \begin{icmlauthorlist}
    \icmlauthor{Ruilin Li}{wu,sii}
    \icmlauthor{Yibin Wang}{sii,fdu}
    \icmlauthor{Wenhong Zhu}{sii,sjtu}
    \icmlauthor{Chenglin Li}{sii} \\
    \icmlauthor{Jinghao Zhang}{sii,ustc}
    \icmlauthor{Chenliang Li}{wu,cor}
    \icmlauthor{Junchi Yan}{sii,sjtu,cor}
    \icmlauthor{Jiaqi Wang}{sii,cor}
  \end{icmlauthorlist}

  \icmlaffiliation{wu}{Wuhan University}
  \icmlaffiliation{sii}{Shanghai Innovation Institute}
  \icmlaffiliation{fdu}{Fudan University}
  \icmlaffiliation{sjtu}{Shanghai Jiao Tong University}
  \icmlaffiliation{ustc}{University of Science and Technology of China}

  \ificmlshowauthors
    \centerline{\texttt{\url{https://github.com/RlinL/EtCon}}}
  \fi

  \vskip 0.3in
]

\printAffiliationsAndNotice{}

\begin{abstract}

Knowledge editing aims to update specific facts in large language models (LLMs) without full retraining.
Prior efforts sought to tune the knowledge layers of LLMs, achieving improved performance in controlled, teacher-forced evaluations.
However, they still encounter challenges in real-world autoregressive generation scenarios, which greatly limit their practical applicability.
Our empirical analysis reveals two issues: \textbf{(1)} Most methods degrade pre-trained capabilities after injecting new knowledge; \textbf{(2)} 
They may exhibit a discrepancy between stored parametric knowledge and inference-time autoregressive generation behavior.
To this end, we propose \textbf{EtCon}, an edit-then-consolidate paradigm that couples targeted edits with post-edit consolidation.
Specifically, our framework comprises two stages: \textbf{(1)} Targeted Proximal Supervised Fine-Tuning (TPSFT) performs a constrained targeted edit to update parametric knowledge while controlling policy drift. \textbf{(2)} Group Relative Policy Optimization (GRPO) consolidates the edit by aligning autoregressive trajectories with the intended fact. 
Extensive experiments demonstrate that our EtCon improves editing reliability and real-world generalization, while better preserving pre-trained capabilities.

\end{abstract}

\section{Introduction}

Large language models (LLMs) have demonstrated unprecedented capabilities across numerous tasks~\cite{wang2025unified,yang2025knowing,he2025llm,liu2025scalecua,rosen2025clinical,shmatko2025learning}.
However, as the world continuously evolves, static pre-trained knowledge quickly becomes outdated~\cite{zheng2025towards}. 
While continual pre-training can refresh knowledge, it is computationally expensive and impractical to perform frequently at scale~\cite{mitchell2022memory}, motivating knowledge editing methods that perform targeted updates through localized parameter modifications~\cite{zhang2025explainable,scialanga2025sake,li2025reinforced,rozner2024knowledge}.
Existing knowledge editing methods roughly fall into two families: 
\textbf{In-Place Parametric Edits}~\cite{meng2022locating,dai2025namet,han2024parameter,fu2025model,jiang2025anyedit} that directly update model weights under the vanilla architecture, and 
\textbf{External-Assisted Edits}~\cite{hartvigsen2023aging,li2025reinforced,tan2023massive,wang2024memoryllm,wang2024wise} that introduce auxiliary modules (e.g., hypernetworks or memory) to support updates at inference time.
Despite encouraging results under single-edit and teacher-forced evaluations, their reliability in real-world autoregressive generation remains limited~\cite{jiang2024learning,tan2023massive,chen2024lifelong,gu2024model,huang2024can}.

\begin{figure*}[t!]
  \centering
  \includegraphics[width=0.9\linewidth]{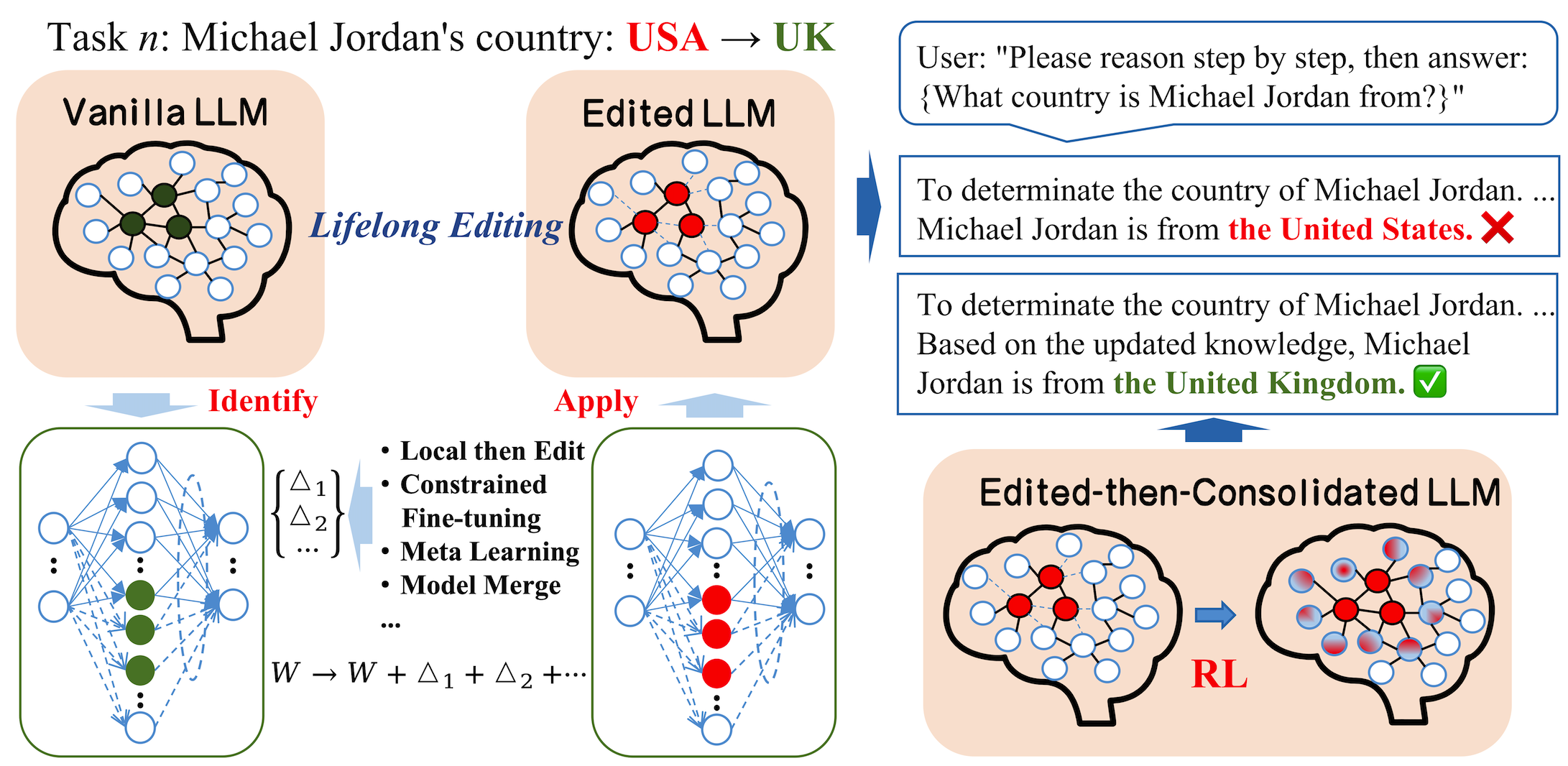}
  \caption{\textbf{Illustration of Knowledge--Behavior Mismatch in Knowledge Editing.} Parametrically stored knowledge may fail to be applied during autoregressive generation.}
  \label{fig:pipe1}
  \vskip -0.15in
\end{figure*}

In this work, our empirical analysis reveals two recurring issues.
\textbf{(1) Pre-trained capability degradation.}
Many existing methods perform edits by maximizing the likelihood of the new target on a small set of editing samples.
Without explicit constraints or regularization, this few-shot objective can induce over-specialization to the editing examples, degrading pre-trained capabilities.
In lifelong editing, as illustrated in Figure~\ref{fig:pipe1} (left), such drift may accumulate across lifelong edits and may even lead to model collapse~\cite{gupta2024rebuilding,gupta2025lifelong}.
\textbf{(2) Parametric knowledge--behavior mismatch in autoregressive generation.}
More critically, Figure~\ref{fig:pipe1} (right) shows that even when the edited fact is stored parametrically, it may not be consistently expressed along autoregressive trajectories.
A key reason is the train-test mismatch between teacher forcing and autoregressive inference: edits are optimized on ground-truth prefixes, whereas real-world generation conditions on self-generated prefixes, inducing a prefix distribution shift that makes edited behavior fragile and prone to reversion.

To address these limitations, we propose \textbf{EtCon}, a new edit-then-consolidate paradigm (Fig.~\ref{fig:pipe}) that couples constrained targeted edits with post-edit consolidation to improve both editing reliability and pre-trained capability preservation.
Our training involves a two-stage pipeline:
\textbf{Stage~1 (Edit).} Targeted
Proximal Supervised Fine-Tuning (TPSFT), a refined variant of PSFT~\cite{zhu2025proximal}, performs a constrained targeted edit by updating only knowledge-relevant FFN layers, with trust-region constraints and self-generated CoT-based targets.
This design localizes parameter updates and mitigates over-specialization, thereby preserving pre-trained capabilities.
\textbf{Stage~2 (Consolidate).} GRPO consolidates the edit via trajectory-level optimization under reward signals.
This stage aligns the parametric update with inference-time behavior, improving the robustness of the edited knowledge along autoregressive trajectories.
We conduct extensive experiments on three datasets with \emph{Llama-3-8B-Instruct} and \emph{Qwen2.5-7B-Instruct}. Under auto-regressive generation with natural stopping and an LLM-as-a-judge protocol \citep{openai_gpt4_1_2025_online}, EtCon improves editing reliability and generalization by 35\%-50\% over strong baselines. It also significantly enhances locality while preserving critical pre-trained capabilities. Our contributions are as follows: 

    (1) We empirically show that incorporating a post-edit consolidation stage improves knowledge editing reliability under realistic scenarios. 

    (2) We propose EtCon, a novel knowledge editing paradigm: TPSFT for localized parametric edits, followed by GRPO for trajectory-level consolidation that aligns parametric knowledge with actual generation behavior.

    (3) Extensive experiments demonstrate that EtCon improves editing reliability and generalization, while preserving pre-trained capabilities under realistic evaluation settings.

\begin{figure*}[t!]
  \centering
  \begin{minipage}[t]{0.48\linewidth}
    \vspace{0pt}
    \centering
    \setlength{\tabcolsep}{0pt}
    \renewcommand{\arraystretch}{1.5}
    {\small
    {\captionsetup{type=table,skip=2pt}\caption{\textbf{Performance Comparison w/ and w/o Knowledge Consolidation under Real-World Evaluation on ZsRE.} (+GRPO) denotes adding our knowledge consolidation stage.}\label{tab:comparison-GRPO}}
    \begin{tabular*}{\linewidth}{@{\extracolsep{\fill}} c l lll}
    \toprule
    & \textbf{Method} & Reli. & Gener. & Local. \\
    \midrule
    \multirow{6}{*}{\rotatebox{90}{Llama-3-8B-Instruct}}
    & Pre-Edit           &  2.8 &  2.4 & 38.6 \\
    & Pre-Edit(+GRPO)    &  5.2 \textcolor{red}{\scriptsize (+2.4)} &  4.7 \textcolor{red}{\scriptsize (+2.3)} & 38.4 \textcolor{blue}{\scriptsize (-0.2)} \\
    & FT-M               & 16.6 & 15.5 & 29.3 \\
    & FT-M(+GRPO)        & 62.9 \textcolor{red}{\scriptsize (+46.3)} & 52.7 \textcolor{red}{\scriptsize (+37.2)} & 24.9 \textcolor{blue}{\scriptsize (-4.4)} \\
    & ALPHAEDIT          & 18.7 & 14.0 &  6.3 \\
    & ALPHAEDIT(+GRPO)   & 50.4 \textcolor{red}{\scriptsize (+31.7)} & 38.7 \textcolor{red}{\scriptsize (+24.7)} &  5.4 \textcolor{blue}{\scriptsize (-0.9)} \\
    \bottomrule
    \end{tabular*}
    }
  \end{minipage}\hfill
  \begin{minipage}[t]{0.43\linewidth}
    \vspace{0pt}
    \centering
    \includegraphics[height=0.22\textheight,keepaspectratio]{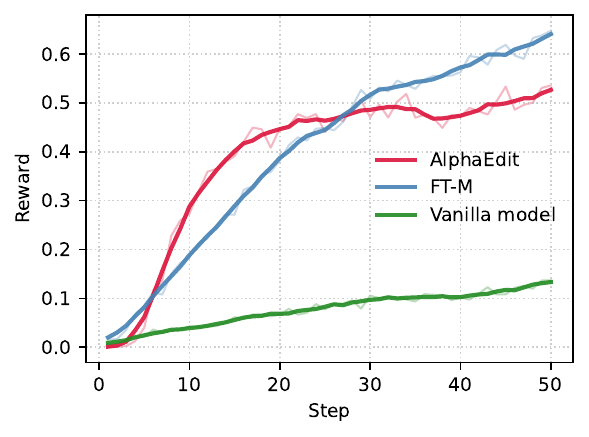}
    {\captionsetup{type=figure,skip=2pt}\caption{\textbf{Reward Curves Comparison.}}\label{fig:reward_curve}}
  \end{minipage}
  \vskip -0.15in
\end{figure*}

\section{Related Work}
\label{sec:related_work}

\subsection{Overview of Knowledge editing methods}

From the perspective of model architecture, the three paradigms introduced above can also be coarsely grouped into two families. \textbf{In-Place Parametric Edits} methods preserve the vanilla LLM architecture. The locate-then-edit paradigm~\cite{meng2022locating,dai2025namet,li2024pmet,zhong2025react,zhang2024locate} identifies knowledge locations within LLMs and modifies targeted parameters through gradient-based or analytical solutions. PEFT methods~\cite{zhu2020modifying,han2024parameter,wang2024deepedit,gupta2025efficient} directly update model parameters via regularized gradient descent to achieve knowledge updates while constraining side effects~\cite{liu2025mitigating}. These approaches seamlessly integrate with existing deployment infrastructure without additional inference latency. \textbf{External-Assisted Edits} methods rely on auxiliary modules for knowledge modification. Meta-learning approaches~\cite{tan2023massive,hartvigsen2023aging,li2025reinforced} train hypernetworks to generate parameter updates, while memory-based methods~\cite{hartvigsen2023aging,zhang2024dafnet,chen2024lifelong} encode knowledge in external modules that the LLM retrieves during inference. Despite their superior performance in balancing reliability and locality, external methods introduce deployment complexity. Given these trade-offs, our work advances In-Place Parametric Edits with a focus on more realistic lifelong knowledge editing scenarios.

\subsection{Evaluation of Knowledge editing methods}

Existing work~\cite{fang2024alphaedit,qi2025incontext,scialanga2025sake} typically evaluates knowledge editing with three standard metrics. 
\textbf{Reliability} measures edit success as the fraction of cases where $P(\mathrm{new~fact}) > P(\mathrm{old~fact})$. 
\textbf{Generalization} applies the same criterion to rephrased queries about the edited knowledge. 
\textbf{Locality} assesses whether the edit preserves responses to related, unedited facts. 
In these protocols, inputs are often templated queries with minimal context, outputs are constrained to a fixed answer format (e.g., via truncation or in-context exemplars), and decoding commonly uses teacher forcing with ground-truth prefixes. 
Recent studies, however, have highlighted the fragility of such evaluation paradigms. 
\textbf{Consequently, this paper adopts a realistic evaluation approach for knowledge editing methods.} 
The full real-world evaluation framework is described in Appendix~\ref{eva}.

\begin{figure*}[t!]
  \centering
  \includegraphics[width=1\linewidth]{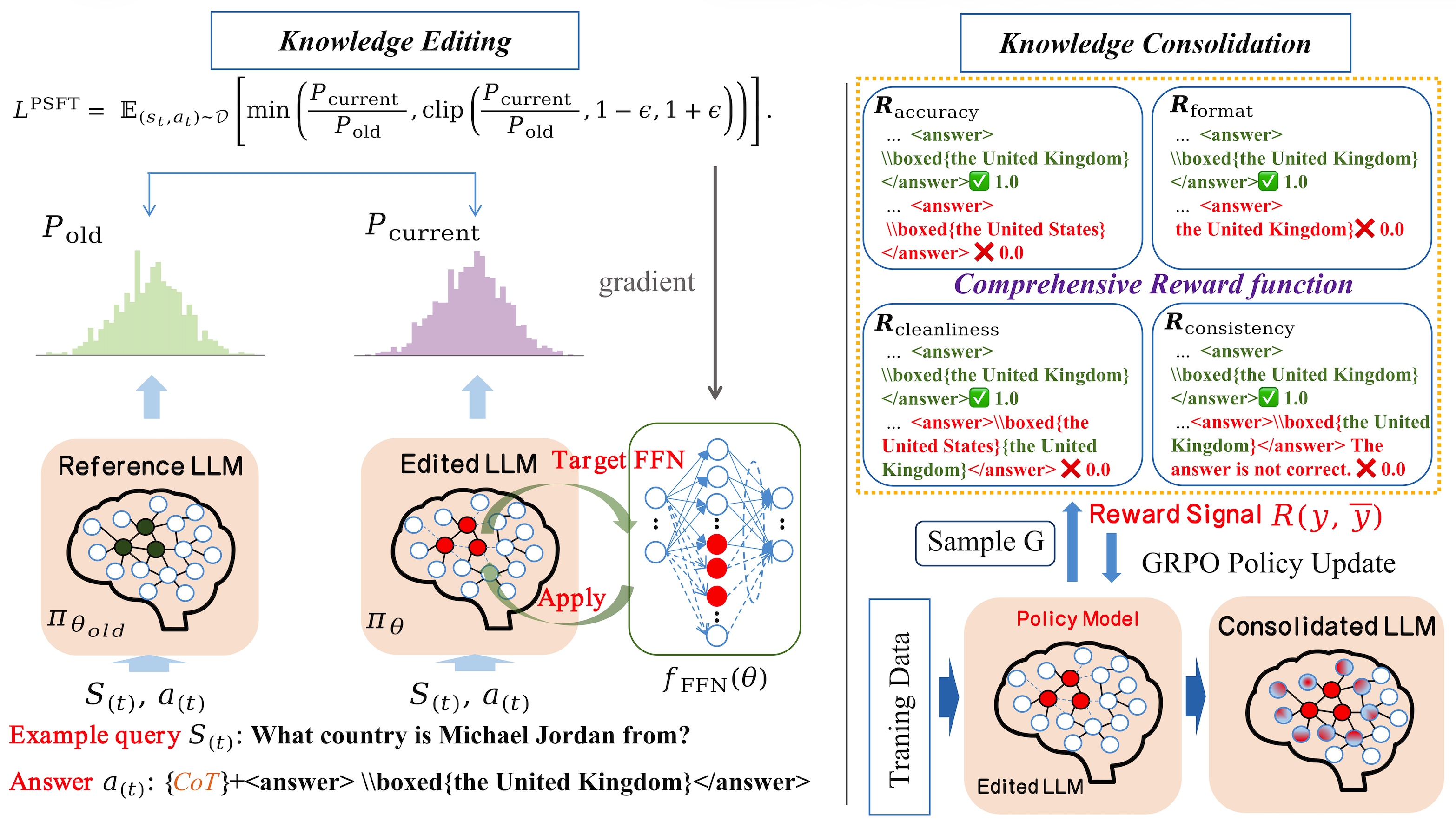}
  \caption{\textbf{Overview of the EtCon Framework.} \textbf{Edit stage} (left): We employ Targeted Proximal Supervised Fine-Tuning (TPSFT) to perform localized edits within the selected FFN layers to inject new knowledge. \textbf{Consolidate stage} (right): We use GRPO with a comprehensive reward function to align the parametric knowledge with actual generation behavior.}
  \label{fig:pipe}
  \vskip -0.15in
\end{figure*}

\section{Preliminary}
\label{Methods}

\subsection{Group Relative Policy Optimization}
\label{subsec:grpo}

Group Relative Policy Optimization (GRPO)~\citep{guo2025deepseek} is a trajectory-level reinforcement learning algorithm that optimizes language model policies through relative reward comparisons within sampled groups. Given a policy $\pi_\theta$ and a dataset $\mathcal{D}$, GRPO maximizes expected reward while constraining deviation from a reference policy $\pi_{\theta_{\text{ref}}}$:
\begin{equation}
\max_{\theta}\;
\mathbb{E}_{(\mathbf{x}, \mathbf{y}^*)\sim\mathcal{D},\;\hat{\mathbf{y}}\sim \pi_\theta(\cdot\mid \mathbf{x})}
\big[r(\mathbf{x}, \mathbf{y}^*, \hat{\mathbf{y}})\big]
- \beta\,D_{\text{KL}}\!\left(\pi_\theta\,\|\,\pi_{\theta_{\text{ref}}}\right),
\end{equation}
where $r(\cdot)$ is a reward function and $\beta$ controls regularization strength. The optimization employs a clipped surrogate objective:
\begin{equation}
\begin{aligned}
J_{\text{GRPO}}(\theta)
&=
\mathbb{E}\!\left[
\sum_{i=1}^{m}
\min\!\Bigl(\rho_i A_i, \text{clip}(\rho_i,\,1-\epsilon,\,1+\epsilon)\,A_i\Bigr)
\right],
\end{aligned}
\end{equation}
where $\rho_i = \pi_\theta(\hat{\mathbf{y}}^i \mid \mathbf{x}) / \pi_{\theta_{\text{rollout}}}(\hat{\mathbf{y}}^i \mid \mathbf{x})$ is the importance ratio, and $A_i = R_i - \frac{1}{m}\sum_{j=1}^{m} R_j$ is the group-relative advantage computed from $m$ trajectories sampled for the same query $\mathbf{x}$. This design eliminates the need for a separate critic network while enabling stable policy updates. Notably, GRPO's trajectory-level optimization makes it naturally suited for aligning knowledge that is parametrically encoded but not reliably applied during autoregressive generation.

\subsection{Motivation for Knowledge Consolidation}
\label{subsec:motivation}

Our key insight is that successful knowledge editing requires a two-stage process rather than parametric knowledge editing alone. Specifically, recent studies reveal a critical gap: while existing methods achieve high success under teacher-forcing evaluation, they can fail in realistic autoregressive generation. Through systematic investigation, we posit that \textbf{an explicit knowledge consolidation stage is crucial for mitigating this gap}. Without a knowledge consolidation stage, edited knowledge may remains superficially encoded at the parametric level, failing to propagate to the model's generation behavior. To validate this hypothesis, we augment existing editing methods with GRPO as a post-editing consolidation step. As shown in Table~\ref{tab:comparison-GRPO} and Figure~\ref{fig:reward_curve}, introducing consolidation dramatically improves both reliability and generalization across multiple baselines. Furthermore, applying GRPO directly to unedited models yields minimal gains, confirming that consolidation requires prior parametric editing as a foundation. These findings establish that \textbf{\textit{parametric knowledge updates and behavioral alignment are complementary but distinct requirements}} for successful knowledge editing.

\section{EtCon}
\label{PSFT}

In this section, we present the EtCon framework (as shown in Figure~\ref{fig:pipe}). 
Stage I employs Targeted Proximal Supervised Fine-Tuning (TPSFT) to perform localized knowledge editing with PPO-style ratio clipping that constrains policy divergence, thereby limiting spillover while preserving pre-trained abilities (\S\ref{subsec:tpsft}); 
Stage II applies GRPO with a comprehensive reward to consolidate edited knowledge at the trajectory level under autoregressive generation (\S\ref{subsec:consolidation}).

\subsection{Knowledge Editing via Targeted Proximal Fine-Tuning}
\label{subsec:tpsft}

In this section, we introduce Targeted Proximal Supervised Fine-Tuning (TPSFT) as a refined knowledge-editing method designed to improve editing efficacy while better preserving pre-trained capabilities.
This approach differs from raw PSFT that updates the whole LLMs, we restrict updates exclusively to specific FFN layers, widely recognized as the primary stores of factual knowledge~\cite{meng2022locating}.
This targeted update strategy effectively injects new knowledge without modifying the model architecture, while minimizing disruption to pre-trained capabilities.

We consider a knowledge editing instance $(\mathbf{x}, \mathbf{y}^*)$, where $\mathbf{x}$ is the input query and $\mathbf{y}^*$ is the target answer. 
Let $\pi_{\theta_{\text{old}}}$ parameterized by $\theta_{\text{old}}$ denote the vanilla LLM before the current edit. 
We partition the \textit{current} parameters $\theta$ into two disjoint sets: the trainable FFN parameters $\theta_{\text{FFN}}$ and the frozen parameters $\theta_{\text{frozen}}$, such that $\theta = \theta_{\text{FFN}} \cup \theta_{\text{frozen}}$. 
The objective of TPSFT is to optimize $\theta_{\text{FFN}}$ to learn the updated parameters $\theta_{\text{new}}$, yielding the post-edit policy $\pi_{\theta_{\text{new}}}$ where $\theta_{\text{new}} = \theta_{\text{FFN}}^{\text{new}} \cup \theta_{\text{frozen}}$.

A distinctive aspect of TPSFT is the use of self-generated CoT-augmented training labels. For each instance $(\mathbf{x}, \mathbf{y}^*)$, we: (1) prompt the vanilla LLM to generate a CoT reasoning path $\mathbf{c}$ for $\mathbf{x}$ (see Appendix~\ref{sec:cot_processing}), and (2) replace the generated answer with $\mathbf{y}^*$, yielding the training target $\mathbf{y} = (\mathbf{c}, \mathbf{y}^*) = (y_1, \dots, y_L)$.
This design integrates the CoT into the supervision signal. Rather than overfitting to a sharp, isolated input-output mapping, the model learns to associate the new knowledge with valid reasoning patterns. To ensure data quality, we filter out generated chains that are logically incompatible with the target (see Appendix~\ref{sec:cot_processing}), thereby preserving the model's intrinsic generation capabilities while injecting the new fact.

Formally, given the editing dataset $\mathcal{D}_r$, for each token position $t$ in the target sequence $\mathbf{y}$, we define the state $\mathbf{S}_t = (\mathbf{x}, \mathbf{y}_{<t})$ and the target token $a_t = y_t$. We optimize $\theta_{\text{FFN}}$ by minimizing the following TPSFT loss:

\begin{equation}
\begin{split}
\mathcal{L}^{\text{TPSFT}}(\theta_{\text{FFN}})
&= - \mathbb{E}_{(\mathbf{S}_t, a_t) \sim \mathcal{D}_r} \Bigl[
\min\bigl(r_t(\theta), \\
&\qquad \text{clip}(r_t(\theta), 1-\epsilon, 1+\epsilon)\bigr)
\Bigr].
\end{split}
\end{equation}

Here, \(\epsilon\) is a hyperparameter that defines the clipping threshold. Following \citet{zhu2025proximal}, the ratio clipping mechanism provides implicit trust-region regularization by (i) bounding the log-probability change per sample to $\log r_t \in [\log(1-\epsilon),\, \log(1+\epsilon)]$, and (ii) nullifying gradients when the ratio exceeds the upper bound.
The probability ratio \(r_t(\theta)\) is the core of this objective and is defined as:
\begin{equation}
r_t(\theta) = \frac{\pi_{\theta}(a_t|\mathbf{S}_t)}{\pi_{\theta_{\text{old}}}(a_t|\mathbf{S}_t)},
\end{equation}
where \(\pi_{\theta}(a_t|\mathbf{S}_t)\) is the probability of generating the target token \(a_t\) given the context \(\mathbf{S}_t\) from the model with \textbf{the current FFN parameters}, and \(\pi_{\theta_{\text{old}}}(a_t|\mathbf{S}_t)\) is the corresponding probability from the \textbf{reference policy}. At the start of the editing process, this reference policy is the initial vanilla LLM. For each subsequent edit instance in the sequential editing process, it is then updated to be the state of the model resulting from the immediately preceding edit.

This objective function acts as a trust-region regularizer via ratio clipping. The term $r_{t}(\theta)$ encourages the model to increase the likelihood of the target answer, similar to standard supervised fine-tuning. Crucially, when this ratio exceeds the threshold $1+\epsilon$, the gradient with respect to $\theta_{\text{FFN}}$ is nullified. This design effectively removes optimization pressure from well-learned samples, constraining parametric deviation and preventing individual edits from dominating the parameter updates.

However, despite these constraints, TPSFT remains a teacher-forced training mechanism implicitly conditioned on ground-truth prefixes $y_{<t}$. During inference, the model must rely on its own generated history $\hat{y}_{<t}$. This discrepancy creates a fundamental \textit{train-inference mismatch}: while the knowledge is parametrically encoded, it may not be reliably elicited during autoregressive generation. This critical limitation necessitates the subsequent consolidation phase.

\subsection{Knowledge Consolidation via GRPO}
\label{subsec:consolidation}

As motivated in \S~\ref{subsec:motivation}, parametric editing alone is insufficient—edited knowledge must be consolidated at the behavioral level. Building on GRPO's trajectory-level optimization (\S~\ref{subsec:grpo}), we design a consolidation algorithm that transforms parametrically encoded knowledge into reliably generated outputs.

Starting from the edited model $\pi_{\theta_{\text{new}}}$ obtained via TPSFT, we optimize the policy $\pi_\theta$ (initialized as $\theta_{\text{new}}$) through autoregressive sampling on the editing dataset $\mathcal{D}_r$. For each query $\mathbf{x}$, the model generates $m$ complete trajectories $\{\hat{\mathbf{y}}^1, \ldots, \hat{\mathbf{y}}^m\}$, which are then scored against the target answer $\mathbf{y}^*$. The group-relative advantage formulation enables the model to learn from comparative trajectory quality without requiring absolute reward calibration. The reference policy is set to $\pi_{\theta_{\text{ref}}}=\pi_{\theta_{\text{new}}}$, ensuring that consolidation refines generation behavior while preserving the parametric edits from Stage I.

\begin{table*}[t!]
\caption{\textbf{Performance Comparison of Lifelong Editing under Real-World Evaluation.} The best results in each group are in \textbf{bold}, and the second-best results are \underline{underlined}. \textcolor{flamingo}{Red numbers} in parentheses indicate the improvement over the second-best method.}
\label{tab:main-results}
\centering

\setlength{\tabcolsep}{2pt} 

\renewcommand{\arraystretch}{1.2}

{\small
\begin{tabularx}{\textwidth}{c l *{3}{>{\hsize=0.85\hsize\raggedleft\arraybackslash}X >{\hsize=0.85\hsize\raggedleft\arraybackslash}X >{\hsize=0.85\hsize\raggedleft\arraybackslash}X >{\hsize=1.45\hsize\centering\arraybackslash}X}}
\toprule
& & \multicolumn{4}{c}{\textbf{ZsRE}} & \multicolumn{4}{c}{\textbf{COUNTERFACT}} & \multicolumn{4}{c}{\textbf{QAEdit}} \\
\cmidrule(lr){3-6} \cmidrule(lr){7-10} \cmidrule(lr){11-14}
& \textbf{Method} & Reli. & Gen. & Loc. & Avg. & Reli. & Gen. & Loc. & Avg. & Reli. & Gen. & Loc. & Avg. \\
\midrule
\multirow{8}{*}{\begin{tabular}{c}Qwen2.5-7B \\-Instruct\end{tabular}}
& Pre-edit &4.4 &3.2 &28.5 & 12.0 &1.0 &0.5 &36.9 & 12.8 &9.8 &10.1 &36.2 & 18.7 \\
\cmidrule(l{\tabcolsep}r{\tabcolsep}){2-14}
& FT-M &5.6 &5.5 &23.1 & 11.4 &\underline{3.2} &\underline{3.1} &24.4 & 10.2 &\underline{14.6} &\underline{14.5} &30.7 & 19.9 \\
& MEMIT     &0.0 &0.1 &0.0 & 0.0 &0.0 &0.2 &0.1 & 0.1 &0.4 &0.3 &0.2 & 0.3 \\
& ALPHAEDIT &\underline{15.9} &\underline{11.5} &6.8 & 11.4 &0.0 &0.0 &0.0 & 0.0 &0.0 &0.0 &0.0 & 0.0 \\
& WISE &4.5 &3.3 &19.1 & 9.0 &1.4 &1.5 &\underline{31.0} & 11.3 &7.1 &9.7 &16.9 & 11.2 \\
& GRACE & 4.8 &4.7 &\textbf{28.1} & \underline{12.5} & 1.9 &1.2 &\textbf{36.8} & \underline{13.3} & 14.2 &13.0 &\textbf{36.0} & \underline{21.1} \\
\rowcolor{lightblue!60} \cellcolor{white} & \textbf{EtCon} & \textbf{69.4} & \textbf{60.8} & \underline{24.4} & \textbf{51.5 {\tiny\textcolor{flamingo}{(+39.0)}}} & \textbf{59.6} & \textbf{43.2} & 29.7 & \textbf{44.2 {\tiny\textcolor{flamingo}{(+30.9)}}} & \textbf{75.1} & \textbf{63.0} & \underline{32.3} & \textbf{56.8 {\tiny\textcolor{flamingo}{(+35.7)}}} \\
\midrule
\multirow{7}{*}{\begin{tabular}{c}Llama-3 \\ -8B-Instruct\end{tabular}}
& Pre-edit &2.8 &2.4 &38.6 & 14.6 &0.6 &0.8 &31.8 & 11.1 &12.7 &12.5 &44.3 & 23.2 \\
\cmidrule(l{\tabcolsep}r{\tabcolsep}){2-14}
& FT-M &16.6 &\underline{15.5} &29.3 & \underline{20.5} &27.9 &18.6 &10.5 & 19.0 &\underline{34.1} &\underline{33.2} &30.1 & \underline{32.5} \\
& MEMIT     &0.1 &0.1 &0.0 & 0.1 &0.3 &0.7 &0.4 & 0.5 &0.2 &0.7 &0.0 & 0.3 \\
& ALPHAEDIT &\underline{18.7} &14.0 &6.3 & 13.0 &\underline{61.0} &\underline{43.8} &16.1 & \underline{40.3} &18.2 &14.9 &7.5 & 13.5 \\
& WISE &4.3 &3.1 &2.2 & 3.2 &1.3 &0.8 &\underline{31.3} & 11.1 &8.1 &13.3 &0.9 & 7.4 \\
& GRACE & 5.5 & 3.9 & \textbf{37.9} & 15.8 &2.1  & 0.7 & \textbf{32.1} & 11.6 & 15.4 & 13.4 & \textbf{34.2} & 21.0 \\
\rowcolor{lightblue!60} \cellcolor{white} & \textbf{EtCon} & \textbf{73.5} & \textbf{63.1} & \underline{30.2} & \textbf{55.6 {\tiny\textcolor{flamingo}{(+35.1)}}} & \textbf{67.1} & \textbf{53.4} & 24.2 & \textbf{48.2 {\tiny\textcolor{flamingo}{(+7.9)}}} & \textbf{70.7} & \textbf{62.7} & \underline{33.6} & \textbf{55.7 {\tiny\textcolor{flamingo}{(+23.2)}}} \\
\bottomrule
\end{tabularx}
}
\end{table*}

The reward function $r_\phi(\mathbf{x}, \mathbf{y}^*, \hat{\mathbf{y}})$ evaluates multiple aspects of the generated trajectory (as illustrated in Figure~\ref{fig:pipe}):
\begin{equation}
\begin{split}
r_\phi(\mathbf{x}, \mathbf{y}^*, \hat{\mathbf{y}}) = w_1 R_{\text{accuracy}} + w_2 R_{\text{format}} \\
\qquad + w_3 R_{\text{cleanliness}} + w_4 R_{\text{consistency}},
\end{split}
\end{equation}
where $R_{\text{accuracy}}$ measures factual accuracy (whether the final answer matches $\mathbf{y}^*$), $R_{\text{format}}$ enforces task-specific output format requirements, $R_{\text{cleanliness}}$ encourages concise outputs without extraneous tokens, and $R_{\text{consistency}}$ rewards internal reasoning coherence and alignment between intermediate steps and the final answer.

This consolidation step integrates the parametric knowledge acquired through TPSFT into the model's inference policy. 
By optimizing trajectory-level generation $\pi_\theta(\hat{\mathbf{y}} \mid \mathbf{x})$ under autoregressive sampling rather than teacher-forcing, GRPO mitigates the train-inference mismatch, ensuring that edited knowledge is reliably elicited in actual behavior.

\section{Experiments}
\label{Exp}

\subsection{Experiment Settings}

\textbf{Datasets and Models.}
This work utilizes 1000 samples from each of four benchmark, ZsRE~\cite{levy2017zero}, COUNTERFACT~\cite{meng2022locating}, MQuAKE-CF-v2~\cite{zhong2023mquake} (1-edit subset), and QAEdit~\cite{yang2025mirage}, to evaluate the performance on knowledge editing tasks. 
We focus on the lifelong editing with sequential edits (batch size 1), which better reflects realistic deployment scenarios where knowledge updates arrive continuously over time. 
We select the Llama-3-8B-Instruct~\cite{dubey2024llama} and Qwen2.5-7B-Instruct~\cite{qwen2025qwen25technicalreport}, as the base models for editing. For general ability evaluation, we use C-Eval~\cite{huang2023c}, CoQA~\cite{reddy2019coqa}, DROP~\cite{dua2019drop}, SQuAD 2.0~\cite{rajpurkar2018know} and LogiQA~\cite{liu2020logiqa}.

\textbf{Baselines.}
We compare our method  against two main categories: In-Place Parametric Edits methods (FT-M~\cite{zhang2024comprehensive}, MEMIT~\cite{meng2022mass}, ALPHAEDIT~\cite{fang2024alphaedit}, MMKE~\cite{fu2025model}) and External-Assisted Edits methods (WISE~\cite{wang2024wise}, GRACE~\cite{hartvigsen2023aging}). In-Place Parametric Edits methods are the main focus of this work, and we select the most representative methods as baselines.

\textbf{Implementation Details.}
We conduct experiments using EasyEdit \cite{xu2025easyedit2} for evaluating various baselines, and employ the lm-evaluation-harness~\cite{eval-harness} for assessing general capabilities. TPSFT is implemented through PSFT~\cite{zhu2025proximal} for edit stage, while GRPO is built upon the EasyR1~\cite{zheng2025easyr1} for the consolidation stage. More details are in Appendix~\ref{e_detail}.

\textbf{Evaluation Metrics.}
We evaluate our method along two principal axes: editing performance and pre-trained capability preservation. To assess editing performance, we employ the LLM-as-judge framework from~\cite{yang2025mirage,gao2024framework,gu2024survey,wang2024lift}, which mitigates the overestimation issue inherent in token-based metrics. Note that while GRPO training uses efficient rule-based rewards (\S\ref{subsec:consolidation}), evaluation employs GPT-4.1 for semantic assessment to ensure rigor.
Specifically, we conduct a binary (correct/incorrect) evaluation of the model's edited outputs. For ZsRE, COUNTERFACT, and QAEdit, we measure Reliability (edit success), Generalization (effectiveness on related inputs), and Locality (impact on unrelated inputs). For MQuAKE-CF-v2, we report Edit-wise accuracy and Multi-hop accuracy. 
 We report pretrained capability on C-Eval and LogiQA, and Exact Match (EM) and F1 on CoQA, DROP, and SQuAD 2.0. Real-world evaluation details are provided in Appendix~\ref{eva}.

\subsection{Main Results}

\textbf{Edit Efficiency.}
EtCon consistently achieves the best overall performance under real-world lifelong editing evaluation across benchmarks and base models.
Table~\ref{tab:main-results} presents  comprehensive results on two LLMs.
Across all six settings (3 benchmarks $\times$ 2 base models), EtCon attains the top \textit{Avg.} score, with large margins over the strongest competing method.
On Qwen2.5-7B-Instruct, EtCon reaches Avg.\ of 51.5/44.2/56.8 on ZsRE/COUNTERFACT/QAEdit (improving the runner-up by +39.0/+30.9/+35.7).
On Llama-3-8B-Instruct, EtCon further achieves Avg.\ of 55.6/48.2/55.7 (+35.1/+7.9/+23.2).
These gains are primarily driven by substantially higher Reliability and Generalization, while maintaining competitive Locality (24.2\%--33.6\%),
indicating that EtCon can internalize new facts and generalize them to paraphrases under sequential edits without overly compromising unrelated behaviors.
We attribute this robustness to our two-stage design: TPSFT performs localized, stability-aware updates, and GRPO subsequently consolidates the edited knowledge under autoregressive trajectories.

\begin{table*}[t!]
\caption{\textbf{Comprehensive Evaluation of Editing Effectiveness and Pre-Trained Capability Preservation.} Comparison on Qwen2.5-7B-Instruct across editing metrics and five general benchmarks.}
\label{tab:com}
\centering
\small
\renewcommand{\arraystretch}{1.25}
\setlength{\tabcolsep}{5pt}
\resizebox{\textwidth}{!}{%
\begin{tabular}{l | p{1.0cm}<{\centering} p{1.0cm}<{\centering} p{1.0cm}<{\centering} p{1.0cm}<{\centering} | p{1.0cm}<{\centering} p{0.7cm}<{\centering} p{0.7cm}<{\centering} p{0.7cm}<{\centering} p{0.7cm}<{\centering} p{0.7cm}<{\centering} p{0.7cm}<{\centering} p{0.85cm}<{\centering} p{0.7cm}<{\centering}}
\toprule
& \multicolumn{4}{c|}{\textit{Editing Performance}} & \multicolumn{9}{c}{\textit{General Capabilities}} \\
\cmidrule(lr){2-5} \cmidrule(l){6-14}
\textbf{Method} & \textbf{Reli.}$\uparrow$ & \textbf{Gen.}$\uparrow$ & \textbf{Loc.}$\uparrow$ & \textbf{Avg.} & \textbf{C-Eval} & \multicolumn{2}{c}{\textbf{CoQA}} & \multicolumn{2}{c}{\textbf{DROP}} & \multicolumn{2}{c}{\textbf{SQuAD}} & \textbf{LogiQA} & \textbf{Avg.} \\
& & & & & Acc. & EM & F1 & EM & F1 & EM & F1 & Acc. & \\
\midrule
Pre & 9.8 & 10.1 & 36.2 & 18.7 & 79.5 & 54.5 & 70.1 & 2.2 & 9.9 & 9.9 & 18.9 & 38.7 & 38.0 \\
\midrule
FT-M & 14.6 & 14.5 & 30.7 & 19.9 & 75.9 & 21.3 & 38.7 & 2.4 & 13.3 & 2.9 & 11.2 & 37.0 & 25.3 \\
FT-M+Con. & \underline{42.3} & \underline{34.1} & 31.9 & \underline{36.1} & 77.0 & 26.2 & 46.6 & 2.8 & 14.6 & 4.4 & 13.5 & 37.8 & 27.9 \\
MMKE & 12.2 & 10.4 & \textbf{34.2} & 18.9 & \textbf{79.3} & \textbf{60.3} & \textbf{74.6} & \textbf{10.3} & \textbf{24.3} & \underline{13.8} & \underline{21.1} & \textbf{41.0} & \textbf{40.6} \\
MMKE+Con. & 37.2 & 31.4 & 31.0 & 33.2 & \underline{78.8} & \underline{59.1} & \underline{73.3} & \underline{8.5} & \underline{21.6} & 12.1 & 19.6 & \underline{39.2} & \underline{39.0} \\
ALPHA & 0.0 & 0.0 & 0.0 & 0.0 & 23.0 & 0.0 & 0.0 & 0.0 & 0.0 & \textbf{50.1} & \textbf{50.1} & 21.8 & 18.1 \\
ALPHA+Con. & 0.0 & 0.0 & 0.0 & 0.0 & 23.0 & 0.0 & 0.0 & 0.0 & 0.0 & \textbf{50.1} & \textbf{50.1} & 21.8 & 18.1 \\
\midrule
\rowcolor{lightblue!60}
\textbf{EtCon} & \textbf{75.1} & \textbf{63.0} & \underline{32.3} & \textbf{56.8} & 78.5 & 55.1 & 69.4 & 2.5 & 8.6 & 9.9 & 19.6 & 38.4 & 35.1 \\
\bottomrule
\end{tabular}%
}
\vskip -0.1in
\end{table*}

\begin{table*}[t]
\centering
\begin{minipage}[t]{0.36\textwidth}
  \vspace{0pt}
  \centering
  {\small
  \captionsetup{font=small}
  \caption{\textbf{Performance Comparison on Multi-Hop Tasks (MQuAKE-CF-v2).} The Base model is evaluated on unedited answers; others on edited answers.}
  \label{tab:editor-cot}
  \vskip 0.03in
  \renewcommand{\arraystretch}{1.0}
  \setlength{\tabcolsep}{5pt}
  \begin{tabular}{lrrrr}
  \toprule
  \textbf{Methods} & \textbf{Edit-wise} & \textbf{2-hop} & \textbf{3-hop} & \textbf{4-hop} \\
  \midrule
  Base & - & 69.1 & 36.2 & 85.4 \\
  \midrule
  FT-M & 19.1 & 2.7 & 3.8 & 4.2 \\
  ALPHA & \underline{20.4} & \underline{13.9} & \underline{11.3} & 9.5 \\
  MMKE & 9.3 & 12.0 & 7.1 & \underline{22.9} \\
  MEMIT & 0.4 & 1.1 & 1.3 & 0.0 \\
  EtCon & \textbf{54.1} & \textbf{24.2} & \textbf{14.6} & \textbf{33.3} \\
  \bottomrule
  \end{tabular}
  }
\end{minipage}
\hfill
\begin{minipage}[t]{0.63\textwidth}
  \vspace{0pt}
  \centering
  \captionsetup{type=figure,font=small}
  \begin{minipage}[t]{0.49\linewidth}
    \vspace{0pt}
    \centering
    \includegraphics[width=\linewidth,height=3.6cm,keepaspectratio]{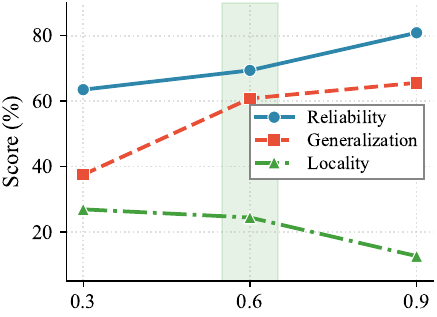}
    \subcaption{Effect of the clipping threshold $\epsilon$}
    \label{fig:param_a}
  \end{minipage}%
  \hspace{2mm}%
  \begin{minipage}[t]{0.49\linewidth}
    \vspace{0pt}
    \centering
    \includegraphics[width=\linewidth,height=3.6cm,keepaspectratio]{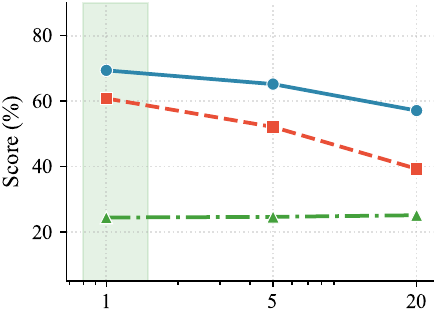}
    \subcaption{Impact of the reference update interval}
    \label{fig:param_b}
  \end{minipage}
  \vskip 0.03in
  \caption{\textbf{Sensitivity Analysis of Key Hyperparameters.}}
  \label{fig:parameter_analysis}
\end{minipage}
\end{table*}

\begin{figure*}[t]
  \centering
  \includegraphics[width=\linewidth]{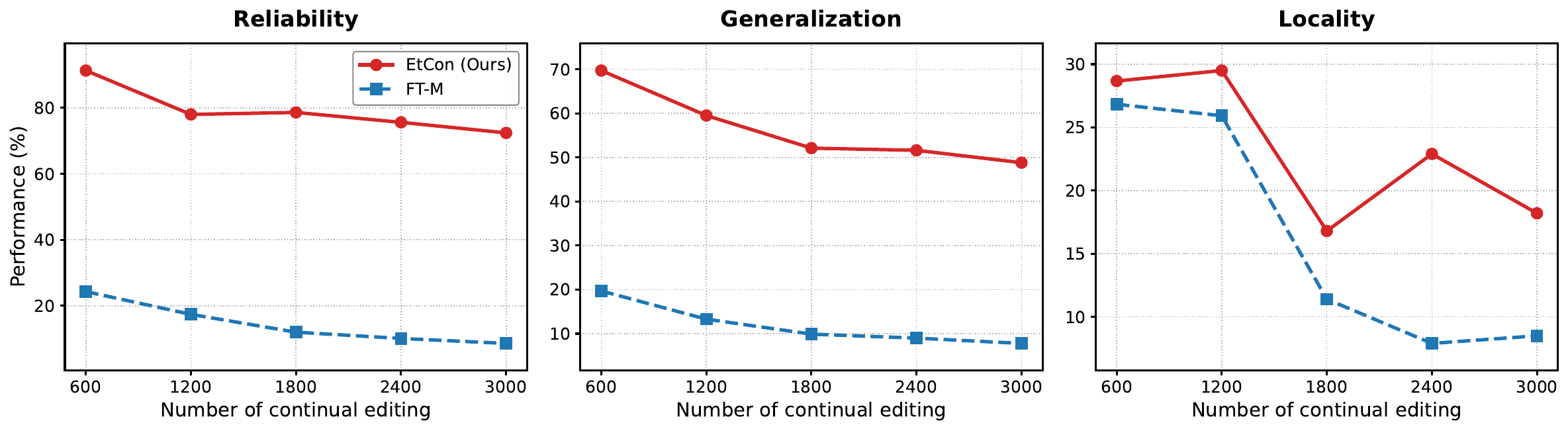}
  \caption{\textbf{Performance Evolution under Larger-Scale Lifelong Editing.}}
  \label{fig:lifelong_edit_compara}
  \vskip -0.1in
\end{figure*}

\begin{table*}[t]
\centering
\begin{minipage}[c]{0.48\textwidth}
    \centering
    \caption{\textbf{Ablation Study of the Key Components in EtCon.} Experiments on COUNTERFACT with Llama-3-8B-Instruct. GC-Avg denotes the mean of general-capability benchmarks (C-Eval (Acc.), CoQA (F1), and SQuAD 2.0 (F1)). Con. denotes Consolidation.}
    \label{tab:ablation-study}
    {\small
    \setlength{\tabcolsep}{2pt}
    \renewcommand{\arraystretch}{0.95}
    \begin{tabular*}{\textwidth}{@{\extracolsep{\fill}} llrrrr}
    \toprule
    \textbf{Stage} & \textbf{Methods} & \textbf{Reli.} & \textbf{Gen.} & \textbf{Loc.} & \textbf{GC-Avg} \\
    \midrule
    Base & - & 0.6 & 0.8 & 31.8 & 52.85 \\
    \midrule
    \multirow{2}{*}{Edit.} 
    &w/ SFT & 1.4 & 0.3 & 30.7  & 50.31 \\
    &w/ TPSFT &3.3  &1.8  & 30.2  & 54.40 \\
    \midrule
    \multirow{3}{*}{\shortstack[l]{Con.\\(after TPSFT)}} 
    & w/o $R_{\text{cleanliness}}$  & 56.1  &22.4  &24.7  & 50.17 \\
    & w/o $R_{\text{consistency}}$  &51.6  & 27.2  &25.1  & 49.75 \\
    & Complete & 67.1  & 53.4  &24.2  & 50.25 \\
    \bottomrule
    \end{tabular*}
    }
\end{minipage}
\hfill
\begin{minipage}[c]{0.48\textwidth}
    \centering
    \includegraphics[width=\linewidth]{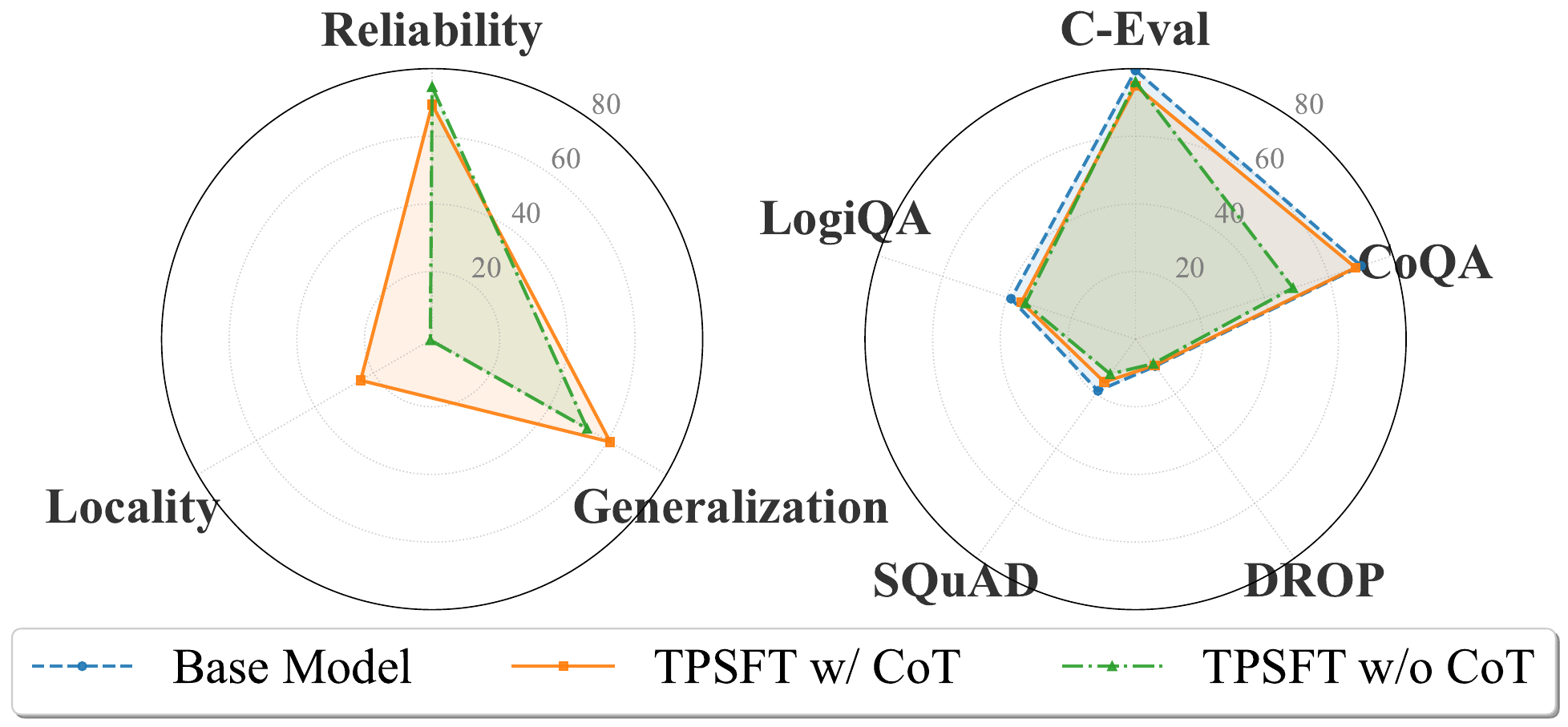}
    \captionof{figure}{\textbf{Effect of CoT Augmentation in EtCon Edit Stage.} Comparison of EtCon with and without CoT-augmented training labels during the TPSFT editing stage.}
    \label{fig:radar-comparison}
\end{minipage}
\end{table*}

\begin{table}[t]
\centering
\caption{\textbf{Comparison of EtCon Edit Scope on QAEdit with Qwen2.5-7B-Instruct.}}
\label{tab:param-update}
{\small
\renewcommand{\arraystretch}{0.95}
\begin{tabular*}{\columnwidth}{@{\extracolsep{\fill}} lrrrr}
\toprule
\textbf{Edit Scope} & \textbf{Reli.} & \textbf{Gen.} & \textbf{Loc.} & \textbf{GC-Avg} \\
\midrule
Full Param.  & 47.3 & 41.5 & 35.8 & 58.17 \\
Target edit & 75.1 & 63.0 & 32.3 & 55.83 \\
\bottomrule
\end{tabular*}
}
\vskip -0.2in
\end{table}

\textbf{Scaling to Longer Editing.}
Figure~\ref{fig:lifelong_edit_compara} shows performance evolution when scaling to 3,000 sequential edits on ZsRE. EtCon maintains stable performance: Reliability and Generalization decrease moderately from 73.5\% to 58.2\% and 63.1\% to 48.7\% respectively, while Locality remains within a narrow band (28\%--32\%). In contrast, FT-M degrades rapidly, with Reliability dropping from 16.6\% to near-zero. These results demonstrate that EtCon remains stable even under longer editing sequences.

\section{Analysis}

\subsection{Effectiveness of Consolidation Stage}

\textbf{Overall Performance of EtCon.}
As shown in Table~\ref{tab:com}, EtCon achieves the best trade-off between editing effectiveness and general capability preservation. EtCon attains 75.1\% Reliability and 63.0\% Generalization, substantially outperforming all baselines. Meanwhile, EtCon maintains general capabilities close to the original model (C-Eval: 78.5 vs 79.5, CoQA F1: 69.4 vs 70.1), demonstrating that our two-stage design successfully balances knowledge injection with pre-trained capability preservation.

\textbf{Performance on Multi-hop Task.} Table~\ref{tab:editor-cot} further demonstrates EtCon's advantage on MQuAKE-CF-v2. EtCon achieves 24.2\% and 33.3\% accuracy on 2-hop and 4-hop queries respectively, substantially outperforming FT-M (2.7\%/4.2\%). This indicates that our consolidation stage can propagate localized edits into more consistent multi-step inference. Under the EtCon paradigm, incorporating explicit reward for multi-hop queries during consolidation could further strengthen performance on multi-hop tasks.

\textbf{Universality of Knowledge Consolidation.}
To evaluate the generality of our knowledge consolidation stage, we augment three baselines (FT-M, MMKE, and ALPHAEDIT) with Consolidation and compare them with EtCon on QAEdit. As shown in Table~\ref{tab:com}, adding Consolidation to FT-M and MMKE improves Reliability by 25\% to 28\% and Generalization by about 20\%. These results suggest that Consolidation bridges the gap between edited parametric knowledge and autoregressive generation behavior, enabling more reliable knowledge utilization in practice.

\textbf{Preservation of Pre-trained Capabilities.}
Evaluations across benchmarks show that the knowledge consolidation stage is non-destructive and preserves general capabilities. However, it cannot recover capabilities lost during editing: once ALPHAEDIT collapses, the knowledge consolidation stage cannot restore it. This indicates that reliable lifelong editing requires both lossless knowledge injection and post-edit consolidation.
Future work under the EtCon paradigm should prioritize lossless knowledge editing.

\subsection{Ablation Study}

\textbf{Edit Stage Ablation.}
We compare TPSFT with standard SFT in Table \ref{tab:ablation-study}. The results show that neither method alone yields reliably editable behavior, as evidenced by consistently low Reliability and Generalization. Nevertheless, TPSFT offers a clear advantage in capability preservation, substantially reducing the degradation observed with SFT.

\textbf{Consolidation Stage Ablation.}
We ablate key components of our consolidation reward (Table~\ref{tab:ablation-study}). 
Removing $R_{\text{cleanliness}}$ causes a clear drop (Reliability $67.1\!\rightarrow\!56.1$, Generalization $53.4\!\rightarrow\!22.4$); inspection shows this encourages ``reward hacking,'' where the model emits extraneous content (e.g., both old and new facts) to inflate scores. 
Removing $R_{\text{consistency}}$ is even more detrimental to Reliability ($67.1\!\rightarrow\!51.6$), often producing catastrophic self-contradictions: the model states the correct answer and then immediately reverts. 
These ablations suggest that both reward components help mitigate reward hacking and improve output coherence; see Appendix~\ref{RH} for case studies.

\textbf{Parameter Sensitivity.} Figure~\ref{fig:parameter_analysis} analyzes TPSFT hyper-parameters.
Clipping threshold $\epsilon$:
Increasing $\epsilon$ improves Reliability and Generalization but reduces Locality, with a sharp drop at $\epsilon=0.9$.
Thus, $\epsilon=0.6$ provides the best trade-off, as larger $\epsilon$ allows stronger updates but increases policy drift.
Reference update interval:
Larger intervals consistently reduce Reliability and Generalization, while Locality remains nearly unchanged.
Frequent updates (Interval$=1$) perform best because stale references increase the policy gap and over-activate clipping, limiting effective updates.

\textbf{Effect of CoT Augmentation.}
Figure \ref{fig:radar-comparison} compares TPSFT with and without CoT-augmented training labels. Without CoT augmentation, Reliability is slightly higher, but Locality drops sharply and general capabilities degrade (e.g., lower CoQA compared to base). This is because supervising only on the few target tokens leads to overfitting. 
CoT augmentation encourages the model to learn valid reasoning patterns rather than overfitting to isolated input-output mappings, thereby better preserving general capabilities.

\textbf{Efficiency of Target Edit.}
We compare \textsc{Target Edit}, which updates a small set of targeted FFN weights, with \textsc{Full Param} update under the same TPSFT setup. As shown in Table~\ref{tab:param-update}, \textsc{Target Edit} yields substantially higher Reliability and Generalization, while incurring only a slight drop in general capability (GC-Avg). In contrast, \textsc{Full Param} attains notably weaker editing performance under the same optimization budget.
We attribute this gap to reduced optimization efficiency under ratio clipping. With limited update steps per edit, \textsc{Full Param} spreads gradients across the whole model, weakening the learning signal on the edited fact. In addition, clipping may saturate early for many tokens, further shrinking effective updates. Overall, restricting TPSFT provides a more controllable and effective mechanism for knowledge editing.

\section{Conclusion}

In this work, we propose \textbf{EtCon}, an edit-then-consolidate framework for reliable knowledge editing.
In Edit Stage, Targeted Proximal Supervised Fine-Tuning (TPSFT), performs constrained edits to inject knowledge while preserving pre-trained capabilities.
In the Consolidate Stage, Group Relative Policy Optimization (GRPO), consolidates edits by aligning autoregressive generation with updated facts.
Extensive experiments demonstrate that EtCon achieves superior editing reliability and generalization while preserving pre-trained capabilities in real-world editing scenarios.

\bibliography{example_paper}

@article{xu2025easyedit2,
  title={EasyEdit2: An Easy-to-use Steering Framework for Editing Large Language Models},
  author={Xu, Ziwen and Wang, Shuxun and Xu, Kewei and Xu, Haoming and Wang, Mengru and Deng, Xinle and Yao, Yunzhi and Zheng, Guozhou and Chen, Huajun and Zhang, Ningyu},
  journal={arXiv preprint arXiv:2504.15133},
  year={2025}
}

@article{wang2025unified,
  title={Unified reward model for multimodal understanding and generation},
  author={Wang, Yibin and Zang, Yuhang and Li, Hao and Jin, Cheng and Wang, Jiaqi},
  journal={arXiv preprint arXiv:2503.05236},
  year={2025}
}

@article{dubey2024llama,
  title={The llama 3 herd of models},
  author={Dubey, Abhimanyu and Jauhri, Abhinav and Pandey, Abhinav and Kadian, Abhishek and Al-Dahle, Ahmad and Letman, Aiesha and Mathur, Akhil and Schelten, Alan and Yang, Amy and Fan, Angela and others},
  journal={arXiv e-prints},
  pages={arXiv--2407},
  year={2024}
}

@article{meng2022mass,
  title={Mass-editing memory in a transformer},
  author={Meng, Kevin and Sharma, Arnab Sen and Andonian, Alex and Belinkov, Yonatan and Bau, David},
  journal={arXiv preprint arXiv:2210.07229},
  year={2022}
}

@article{fang2024alphaedit,
  title={Alphaedit: Null-space constrained knowledge editing for language models},
  author={Fang, Junfeng and Jiang, Houcheng and Wang, Kun and Ma, Yunshan and Jie, Shi and Wang, Xiang and He, Xiangnan and Chua, Tat-Seng},
  journal={arXiv preprint arXiv:2410.02355},
  year={2024}
}

@article{huang2023c,
  title={C-eval: A multi-level multi-discipline chinese evaluation suite for foundation models},
  author={Huang, Yuzhen and Bai, Yuzhuo and Zhu, Zhihao and Zhang, Junlei and Zhang, Jinghan and Su, Tangjun and Liu, Junteng and Lv, Chuancheng and Zhang, Yikai and Fu, Yao and others},
  journal={Advances in Neural Information Processing Systems},
  volume={36},
  pages={62991--63010},
  year={2023}
}

@article{reddy2019coqa,
  title={Coqa: A conversational question answering challenge},
  author={Reddy, Siva and Chen, Danqi and Manning, Christopher D},
  journal={Transactions of the Association for Computational Linguistics},
  volume={7},
  pages={249--266},
  year={2019},
  publisher={MIT Press One Rogers Street, Cambridge, MA 02142-1209, USA journals-info~…}
}

@article{dua2019drop,
  title={DROP: A reading comprehension benchmark requiring discrete reasoning over paragraphs},
  author={Dua, Dheeru and Wang, Yizhong and Dasigi, Pradeep and Stanovsky, Gabriel and Singh, Sameer and Gardner, Matt},
  journal={arXiv preprint arXiv:1903.00161},
  year={2019}
}

@article{wang2024lift,
  title={Lift: Leveraging human feedback for text-to-video model alignment},
  author={Wang, Yibin and Tan, Zhiyu and Wang, Junyan and Yang, Xiaomeng and Jin, Cheng and Li, Hao},
  journal={arXiv preprint arXiv:2412.04814},
  year={2024}
}

@article{rajpurkar2018know,
  title={Know what you don't know: Unanswerable questions for SQuAD},
  author={Rajpurkar, Pranav and Jia, Robin and Liang, Percy},
  journal={arXiv preprint arXiv:1806.03822},
  year={2018}
}

@article{liu2020logiqa,
  title={Logiqa: A challenge dataset for machine reading comprehension with logical reasoning},
  author={Liu, Jian and Cui, Leyang and Liu, Hanmeng and Huang, Dandan and Wang, Yile and Zhang, Yue},
  journal={arXiv preprint arXiv:2007.08124},
  year={2020}
}

@article{zhu2025proximal,
  title={Proximal supervised fine-tuning},
  author={Zhu, Wenhong and Xie, Ruobing and Wang, Rui and Sun, Xingwu and Wang, Di and Liu, Pengfei},
  journal={arXiv preprint arXiv:2508.17784},
  year={2025}
}

@misc{zheng2025easyr1,
  title        = {EasyR1: An Efficient, Scalable, Multi-Modality RL Training Framework},
  author       = {Zheng, Yaowei and Lu, Junting and Wang, Shenzhi and Feng, Zhangchi and Kuang, Dongdong and Xiong, Yuwen },
  howpublished = {\url{https://github.com/hiyouga/EasyR1}},
  year         = {2025}
}

@article{fu2025model,
  title={Model Merging for Knowledge Editing},
  author={Fu, Zichuan and Wu, Xian and Li, Guojing and Zhang, Yingying and Zheng, Yefeng and Ming, Tianshi and Wang, Yejing and Wang, Wanyu and Zhao, Xiangyu},
  journal={arXiv preprint arXiv:2506.12384},
  year={2025}
}

@article{levy2017zero,
  title={Zero-shot relation extraction via reading comprehension},
  author={Levy, Omer and Seo, Minjoon and Choi, Eunsol and Zettlemoyer, Luke},
  journal={arXiv preprint arXiv:1706.04115},
  year={2017}
}

@article{wang2024wise,
  title={Wise: Rethinking the knowledge memory for lifelong model editing of large language models},
  author={Wang, Peng and Li, Zexi and Zhang, Ningyu and Xu, Ziwen and Yao, Yunzhi and Jiang, Yong and Xie, Pengjun and Huang, Fei and Chen, Huajun},
  journal={Advances in Neural Information Processing Systems},
  volume={37},
  pages={53764--53797},
  year={2024}
}

@article{yang2025mirage,
  title={The mirage of model editing: Revisiting evaluation in the wild},
  author={Yang, Wanli and Sun, Fei and Tan, Jiajun and Ma, Xinyu and Cao, Qi and Yin, Dawei and Shen, Huawei and Cheng, Xueqi},
  journal={arXiv preprint arXiv:2502.11177},
  year={2025}
}

@article{rosen2025clinical,
  title={Clinical implementation of an AI-based prediction model for decision support for patients undergoing colorectal cancer surgery},
  author={Rosen, Andreas Weinberger and Ose, Ilze and G{\"o}genur, Mikail and Andersen, Lars Peter Kloster and Bojesen, Rasmus Dahlin and Vogelsang, Rasmus Peuliche and Rose, Martin H{\o}yer and Steenfos, Philip Wallentin and Hansen, Lasse Bremholm and Spuur, Helle Skadborg and others},
  journal={Nature Medicine},
  pages={1--12},
  year={2025},
  publisher={Nature Publishing Group US New York}
}

@article{he2025llm,
  title={LLM-Based Multi-Agent Systems for Software Engineering: Literature Review, Vision, and the Road Ahead},
  author={He, Junda and Treude, Christoph and Lo, David},
  journal={ACM Transactions on Software Engineering and Methodology},
  volume={34},
  number={5},
  pages={1--30},
  year={2025},
  publisher={ACM New York, NY}
}

@article{shmatko2025learning,
  title={Learning the natural history of human disease with generative transformers},
  author={Shmatko, Artem and Jung, Alexander Wolfgang and Gaurav, Kumar and Brunak, S{\o}ren and Mortensen, Laust Hvas and Birney, Ewan and Fitzgerald, Tom and Gerstung, Moritz},
  journal={Nature},
  pages={1--9},
  year={2025},
  publisher={Nature Publishing Group UK London}
}

@article{liu2025scalecua,
  title={ScaleCUA: Scaling Open-Source Computer Use Agents with Cross-Platform Data},
  author={Liu, Zhaoyang and Xie, JingJing and Ding, Zichen and Li, Zehao and Yang, Bowen and Wu, Zhenyu and Wang, Xuehui and Sun, Qiushi and Liu, Shi and Wang, Weiyun and others},
  journal={arXiv preprint arXiv:2509.15221},
  year={2025}
}

@inproceedings{yang2025knowing,
  title={Knowing You Don't Know: Learning When to Continue Search in Multi-round RAG through Self-Practicing},
  author={Yang, Diji and Zeng, Linda and Rao, Jinmeng and Zhang, Yi},
  booktitle={Proceedings of the 48th International ACM SIGIR Conference on Research and Development in Information Retrieval},
  pages={1305--1315},
  year={2025}
}

@article{meng2022locating,
  title={Locating and editing factual associations in gpt},
  author={Meng, Kevin and Bau, David and Andonian, Alex and Belinkov, Yonatan},
  journal={Advances in neural information processing systems},
  volume={35},
  pages={17359--17372},
  year={2022}
}

@article{han2024parameter,
  title={Parameter-efficient fine-tuning for large models: A comprehensive survey},
  author={Han, Zeyu and Gao, Chao and Liu, Jinyang and Zhang, Jeff and Zhang, Sai Qian},
  journal={arXiv preprint arXiv:2403.14608},
  year={2024}
}

@article{guo2025deepseek,
  title={Deepseek-r1 incentivizes reasoning in llms through reinforcement learning},
  author={Guo, Daya and Yang, Dejian and Zhang, Haowei and Song, Junxiao and Wang, Peiyi and Zhu, Qihao and Xu, Runxin and Zhang, Ruoyu and Ma, Shirong and Bi, Xiao and others},
  journal={Nature},
  volume={645},
  number={8081},
  pages={633--638},
  year={2025},
  publisher={Nature Publishing Group UK London}
}

@inproceedings{mitchell2022memory,
  title={Memory-based model editing at scale},
  author={Mitchell, Eric and Lin, Charles and Bosselut, Antoine and Manning, Christopher D and Finn, Chelsea},
  booktitle={International Conference on Machine Learning},
  pages={15817--15831},
  year={2022},
  organization={PMLR}
}

@inproceedings{zhang2025explainable,
  title={Explainable and efficient editing for large language models},
  author={Zhang, Tianyu and Fang, Junfeng and Jiang, Houcheng and Bi, Baolong and Wang, Xiang and He, Xiangnan},
  booktitle={Proceedings of the ACM on Web Conference 2025},
  pages={1963--1976},
  year={2025}
}

@article{zheng2025towards,
  title={Towards lifelong learning of large language models: A survey},
  author={Zheng, Junhao and Qiu, Shengjie and Shi, Chengming and Ma, Qianli},
  journal={ACM Computing Surveys},
  volume={57},
  number={8},
  pages={1--35},
  year={2025},
  publisher={ACM New York, NY}
}

@article{scialanga2025sake,
  title={SAKE: Steering Activations for Knowledge Editing},
  author={Scialanga, Marco and Laugel, Thibault and Grari, Vincent and Detyniecki, Marcin},
  journal={arXiv preprint arXiv:2503.01751},
  year={2025}
}

@article{rozner2024knowledge,
  title={Knowledge editing in language models via adapted direct preference optimization},
  author={Rozner, Amit and Battash, Barak and Wolf, Lior and Lindenbaum, Ofir},
  journal={arXiv preprint arXiv:2406.09920},
  year={2024}
}

@article{wang2024memoryllm,
  title={Memoryllm: Towards self-updatable large language models},
  author={Wang, Yu and Gao, Yifan and Chen, Xiusi and Jiang, Haoming and Li, Shiyang and Yang, Jingfeng and Yin, Qingyu and Li, Zheng and Li, Xian and Yin, Bing and others},
  journal={arXiv preprint arXiv:2402.04624},
  year={2024}
}

@article{jiang2024learning,
  title={Learning to edit: Aligning llms with knowledge editing},
  author={Jiang, Yuxin and Wang, Yufei and Wu, Chuhan and Zhong, Wanjun and Zeng, Xingshan and Gao, Jiahui and Li, Liangyou and Jiang, Xin and Shang, Lifeng and Tang, Ruiming and others},
  journal={arXiv preprint arXiv:2402.11905},
  year={2024}
}

@article{tan2023massive,
  title={Massive editing for large language models via meta learning},
  author={Tan, Chenmien and Zhang, Ge and Fu, Jie},
  journal={arXiv preprint arXiv:2311.04661},
  year={2023}
}

@article{gu2024model,
  title={Model editing harms general abilities of large language models: Regularization to the rescue},
  author={Gu, Jia-Chen and Xu, Hao-Xiang and Ma, Jun-Yu and Lu, Pan and Ling, Zhen-Hua and Chang, Kai-Wei and Peng, Nanyun},
  journal={arXiv preprint arXiv:2401.04700},
  year={2024}
}

@article{huang2024can,
  title={Can Knowledge Editing Really Correct Hallucinations?},
  author={Huang, Baixiang and Chen, Canyu and Xu, Xiongxiao and Payani, Ali and Shu, Kai},
  journal={arXiv preprint arXiv:2410.16251},
  year={2024}
}

@article{wang2024deepedit,
  title={Deepedit: Knowledge editing as decoding with constraints},
  author={Wang, Yiwei and Chen, Muhao and Peng, Nanyun and Chang, Kai-Wei},
  journal={arXiv preprint arXiv:2401.10471},
  year={2024}
}

@article{gupta2025efficient,
  title={Efficient Knowledge Editing via Minimal Precomputation},
  author={Gupta, Akshat and Lu, Maochuan and Hartvigsen, Thomas and Anumanchipalli, Gopala},
  journal={arXiv preprint arXiv:2506.04226},
  year={2025}
}

@article{zhu2020modifying,
  title={Modifying memories in transformer models},
  author={Zhu, Chen and Rawat, Ankit Singh and Zaheer, Manzil and Bhojanapalli, Srinadh and Li, Daliang and Yu, Felix and Kumar, Sanjiv},
  journal={arXiv preprint arXiv:2012.00363},
  year={2020}
}

@article{dai2025namet,
  title={NAMET: Robust Massive Model Editing via Noise-Aware Memory Optimization},
  author={Dai, Yanbo and Ji, Zhenlan and Li, Zongjie and Wang, Shuai},
  journal={arXiv preprint arXiv:2505.11876},
  year={2025}
}

@inproceedings{li2024pmet,
  title={Pmet: Precise model editing in a transformer},
  author={Li, Xiaopeng and Li, Shasha and Song, Shezheng and Yang, Jing and Ma, Jun and Yu, Jie},
  booktitle={Proceedings of the AAAI Conference on Artificial Intelligence},
  pages={18564--18572},
  year={2024}
}

@article{gu2024survey,
  title={A survey on llm-as-a-judge},
  author={Gu, Jiawei and Jiang, Xuhui and Shi, Zhichao and Tan, Hexiang and Zhai, Xuehao and Xu, Chengjin and Li, Wei and Shen, Yinghan and Ma, Shengjie and Liu, Honghao and others},
  journal={arXiv preprint arXiv:2411.15594},
  year={2024}
}

@misc{gao2024framework,
      title={A framework for few-shot language model evaluation}, 
      author={Leo Gao and Jonathan Tow and Baber Abbasi and Stella Biderman and Sid Black and Anthony DiPofi and Charles Foster and Laurence Golding and Jeffrey Hsu and Alain Le Noac'h and Haonan Li and Kyle McDonell and Niklas Muennighoff and Chris Ociepa and Jason Phang and Laria Reynolds and Hailey Schoelkopf and Aviya Skowron and Lintang Sutawika and Eric Tang and Anish Thite and Ben Wang and Kevin Wang and Andy Zou},
      year={2024},
      eprint={10.5281/zenodo.12608602},
      archivePrefix={Zenodo},
      doi={10.5281/zenodo.12608602}
}

@misc{eval-harness,
  author       = {Gao, Leo and Tow, Jonathan and Abbasi, Baber and Biderman, Stella and Black, Sid and DiPofi, Anthony and Foster, Charles and Golding, Laurence and Hsu, Jeffrey and Le Noac'h, Alain and Li, Haonan and McDonell, Kyle and Muennighoff, Niklas and Ociepa, Chris and Phang, Jason and Reynolds, Laria and Schoelkopf, Hailey and Skowron, Aviya and Sutawika, Lintang and Tang, Eric and Thite, Anish and Wang, Ben and Wang, Kevin and Zou, Andy},
  title        = {The Language Model Evaluation Harness},
  month        = 07,
  year         = 2024,
  publisher    = {Zenodo},
  version      = {v0.4.3},
  doi          = {10.5281/zenodo.12608602},
  url          = {https://zenodo.org/records/12608602}
}

@inproceedings{
qi2025incontext,
title={In-Context Editing: Learning Knowledge from Self-Induced Distributions},
author={Siyuan Qi and Bangcheng Yang and Kailin Jiang and Xiaobo Wang and Jiaqi Li and Yifan Zhong and Yaodong Yang and Zilong Zheng},
booktitle={The Thirteenth International Conference on Learning Representations},
year={2025},
url={https://openreview.net/forum?id=w6rHCuN3YG}
}

@article{zhong2025react,
  title={REACT: Representation Extraction And Controllable Tuning to Overcome Overfitting in LLM Knowledge Editing},
  author={Zhong, Haitian and Liu, Yuhuan and Xu, Ziyang and Liu, Guofan and Liu, Qiang and Wu, Shu and Zhao, Zhe and Wang, Liang and Tan, Tieniu},
  journal={arXiv preprint arXiv:2505.18933},
  year={2025}
}

@article{hartvigsen2023aging,
  title={Aging with grace: Lifelong model editing with discrete key-value adaptors},
  author={Hartvigsen, Tom and Sankaranarayanan, Swami and Palangi, Hamid and Kim, Yoon and Ghassemi, Marzyeh},
  journal={Advances in Neural Information Processing Systems},
  volume={36},
  pages={47934--47959},
  year={2023}
}

@article{li2025reinforced,
  title={Reinforced lifelong editing for language models},
  author={Li, Zherui and Jiang, Houcheng and Chen, Hao and Bi, Baolong and Zhou, Zhenhong and Sun, Fei and Fang, Junfeng and Wang, Xiang},
  journal={arXiv preprint arXiv:2502.05759},
  year={2025}
}

@article{gupta2024rebuilding,
  title={Rebuilding rome: Resolving model collapse during sequential model editing},
  author={Gupta, Akshat and Baskaran, Sidharth and Anumanchipalli, Gopala},
  journal={arXiv preprint arXiv:2403.07175},
  year={2024}
}

@article{gupta2025lifelong,
  title={Lifelong Sequential Knowledge Editing without Model Degradation},
  author={Gupta, Akshat and Prateepamornkul, Phudish and Lu, Maochuan and Alaa, Ahmed and Hartvigsen, Thomas and Anumanchipalli, Gopala},
  journal={arXiv e-prints},
  pages={arXiv--2502},
  year={2025}
}

@article{zhang2024dafnet,
  title={Dafnet: Dynamic auxiliary fusion for sequential model editing in large language models},
  author={Zhang, Taolin and Chen, Qizhou and Li, Dongyang and Wang, Chengyu and He, Xiaofeng and Huang, Longtao and Xue, Hui and Huang, Jun},
  journal={arXiv preprint arXiv:2405.20588},
  year={2024}
}

@article{zhang2024locate,
  title={Locate-then-edit for multi-hop factual recall under knowledge editing},
  author={Zhang, Zhuoran and Li, Yongxiang and Kan, Zijian and Cheng, Keyuan and Hu, Lijie and Wang, Di},
  journal={arXiv preprint arXiv:2410.06331},
  year={2024}
}

@article{liu2025mitigating,
  title={Mitigating heterogeneous token overfitting in llm knowledge editing},
  author={Liu, Tianci and Li, Ruirui and Dong, Zihan and Liu, Hui and Tang, Xianfeng and Yin, Qingyu and Zhang, Linjun and Wang, Haoyu and Gao, Jing},
  journal={arXiv preprint arXiv:2502.00602},
  year={2025}
}

@article{chen2024lifelong,
  title={Lifelong knowledge editing for llms with retrieval-augmented continuous prompt learning},
  author={Chen, Qizhou and Zhang, Taolin and He, Xiaofeng and Li, Dongyang and Wang, Chengyu and Huang, Longtao and Xue, Hui},
  journal={arXiv preprint arXiv:2405.03279},
  year={2024}
}

@online{openai_gpt4_1_2025_online,
  author={OpenAI},
  title={Introducing {GPT}-4.1 in the {API}},
  url={https://openai.com/index/gpt-4-1/},
  year={2025}
}

@article{zhang2024comprehensive,
  title={A comprehensive study of knowledge editing for large language models},
  author={Zhang, Ningyu and Yao, Yunzhi and Tian, Bozhong and Wang, Peng and Deng, Shumin and Wang, Mengru and Xi, Zekun and Mao, Shengyu and Zhang, Jintian and Ni, Yuansheng and others},
  journal={arXiv preprint arXiv:2401.01286},
  year={2024}
}

@inproceedings{geva2021transformer,
  title={Transformer feed-forward layers are key-value memories},
  author={Geva, Mor and Schuster, Roei and Berant, Jonathan and Levy, Omer},
  booktitle={Proceedings of the 2021 Conference on Empirical Methods in Natural Language Processing},
  pages={5484--5495},
  year={2021}
}

@inproceedings{qicontext,
  title={In-Context Editing: Learning Knowledge from Self-Induced Distributions},
  author={Qi, Siyuan and Yang, Bangcheng and Jiang, Kailin and Wang, Xiaobo and Li, Jiaqi and Zhong, Yifan and Yang, Yaodong and Zheng, Zilong},
  booktitle={The Thirteenth International Conference on Learning Representations},
  year={2024}
}

@article{jiang2025anyedit,
  title={Anyedit: Edit any knowledge encoded in language models},
  author={Jiang, Houcheng and Fang, Junfeng and Zhang, Ningyu and Ma, Guojun and Wan, Mingyang and Wang, Xiang and He, Xiangnan and Chua, Tat-seng},
  journal={arXiv preprint arXiv:2502.05628},
  year={2025}
}

@article{zhong2023mquake,
  title={Mquake: Assessing knowledge editing in language models via multi-hop questions},
  author={Zhong, Zexuan and Wu, Zhengxuan and Manning, Christopher D and Potts, Christopher and Chen, Danqi},
  journal={arXiv preprint arXiv:2305.14795},
  year={2023}
}

@misc{qwen2025qwen25technicalreport,
      title={Qwen2.5 Technical Report}, 
      author={Qwen and : and An Yang and Baosong Yang and Beichen Zhang and Binyuan Hui and Bo Zheng and Bowen Yu and Chengyuan Li and Dayiheng Liu and Fei Huang and Haoran Wei and Huan Lin and Jian Yang and Jianhong Tu and Jianwei Zhang and Jianxin Yang and Jiaxi Yang and Jingren Zhou and Junyang Lin and Kai Dang and Keming Lu and Keqin Bao and Kexin Yang and Le Yu and Mei Li and Mingfeng Xue and Pei Zhang and Qin Zhu and Rui Men and Runji Lin and Tianhao Li and Tianyi Tang and Tingyu Xia and Xingzhang Ren and Xuancheng Ren and Yang Fan and Yang Su and Yichang Zhang and Yu Wan and Yuqiong Liu and Zeyu Cui and Zhenru Zhang and Zihan Qiu},
      year={2025},
      eprint={2412.15115},
      archivePrefix={arXiv},
      primaryClass={cs.CL},
      url={https://arxiv.org/abs/2412.15115}, 
}
\bibliographystyle{icml2026}

\clearpage
\onecolumn
\appendix

\section{Appendix}
\subsection{Additional Implementation Details}
\label{e_detail}

\textbf{Baseline Configuration.}
For our baseline experiments, we utilize the EasyEdit framework. All hyperparameters adhere to the default configurations of the respective comparison methods, with further details provided in \cite{yang2025mirage, qi2025incontext}. 

\textbf{Editing Stage.}
For our proposed EtCon method, we update only the FFN down-projection layers (mlp.down\_proj) in layers 7–11 of Llama-3-8B-Instruct and layers 5–9 of Qwen2.5-7B-Instruct, following prior works~\cite{zhang2024comprehensive, meng2022locating, geva2021transformer, qicontext} on knowledge editing.
Following the lifelong editing paradigm, we use AdamW with learning rate \(1\times10^{-5}\) and set \(\epsilon=0.6\) of TPSFT, sequentially performing edits (batch size 1) on the same model. TPSFT is trained for 5 epochs with at most 6 update steps per edit using early stopping. 

\textbf{Consolidation Stage.}
We optimize the inference-time policy with GRPO. The comprehensive reward function in Equation (5) uses the following weight coefficients: $w_1 = 0.7$ for $R_{\text{accuracy}}$, $w_2 = 0.05$ for $R_{\text{format}}$, $w_3 = 0.15$ for $R_{\text{cleanliness}}$, and $w_4 = 0.1$ for $R_{\text{consistency}}$. These weights were determined through extensive empirical experiments to balance factual accuracy with output quality. All reward components are binary (0/1), with $R_{\text{accuracy}}$ computed via exact string matching between the extracted answer and the ground-truth after normalization. All specific hyperparameters are available in Table \ref{tab:training_config}.

\begin{table}[H]
\vspace{-1em}
\centering
\caption{\textbf{Hyperparameters for the GRPO Consolidation Stage.}}
\label{tab:training_config}
{\setlength{\tabcolsep}{4pt}
\begin{tabular}{@{}l l@{}}
\toprule
\textbf{Configuration} & \textbf{Value} \\
\midrule
\multicolumn{2}{l}{\textit{Model Configuration}} \\
Precision & BFloat16 \\
Max Prompt Length & 2k \\
Max Response Length & 2k \\
\midrule
\multicolumn{2}{l}{\textit{Training Hyperparameters}} \\
Learning Rate & $1.0 \times 10^{-6}$ \\
Optimizer & AdamW (BF16 variant) \\
Global Batch Size & 64 \\
Rollout Batch Size & 256 \\
Micro Batch Size (Update) & 4 \\
Micro Batch Size (Experience) & 16 \\
Training Step & 100 \\
Gradient Clipping & 1.0 \\
\midrule
\multicolumn{2}{l}{\textit{Rollout Configuration}} \\
Number of Rollouts ($m$) & 8 \\
Temperature & 1.0 \\
Top-p & 0.99 \\
\midrule
\multicolumn{2}{l}{\textit{Infrastructure}} \\
GPUs & 8 × NVIDIA H800 \\
Tensor Parallelism & 1 \\
FSDP & Enabled \\
CPU Offloading & Disabled \\
Gradient Checkpointing & Enabled \\
\midrule
\multicolumn{2}{l}{\textit{Validation}} \\
Validation Batch Size & 512 \\
Validation Frequency & Every 5 steps \\
Validation before Training & Yes \\
\bottomrule
\end{tabular}
}
\end{table}

\newpage
\subsection{Additional Results}

\subsubsection{Effect of Edited Layers on EtCon}
\label{diff_layer_appendix}

\begin{figure}[H]
\centering
\begin{minipage}[t]{0.38\linewidth}
  \vspace{0pt}
  \centering
  \captionof{table}{\textbf{Layer Selection Ablation on Llama-3-8B-Instruct.}}
  \label{tab:diff_layer}
  \vspace{2pt}
  \setlength{\tabcolsep}{5pt}
  \renewcommand{\arraystretch}{1.15}
  {\large
  \begin{tabular}{l rrr}
    \toprule
    \textbf{Layers} & Reli. & Gener. & Local. \\
    \midrule
    7-8-9-10-11      & 73.5 & 63.1 & 30.2 \\
    12-13-14-15-16   & 78.1 & 58.3 & 24.1 \\
    17-18-19-20-21   & 76.7 & 53.2 & 17.3 \\
    \bottomrule
  \end{tabular}
  }
\end{minipage}\hfill
\begin{minipage}[t]{0.5\linewidth}
  \vspace{0pt}
  \centering
  \includegraphics[width=\linewidth]{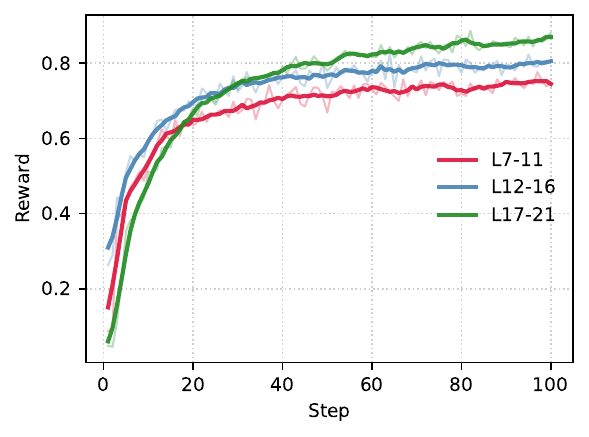}
  \captionof{figure}{\textbf{Reward Convergence across Layer Configurations.}}
  \label{fig:diff_layer_reward_curve}
\end{minipage}
\vskip -0.2in
\end{figure}

\textbf{Experimental Setup.}
We further investigate the impact of editing different FFN layers on EtCon's performance through an ablation study, as shown in Table \ref{tab:diff_layer} and Figure \ref{fig:diff_layer_reward_curve}.

\textbf{Main Finding.}
We observe that under identical hyperparameter settings, editing early layers (Layers 7-11) outperforms deeper layers (Layers 17-21) in both locality and generalization.

\textbf{Reward Hacking Phenomenon.}
Analyzing the ``high reward, low performance'' phenomenon in deeper layers shown in Figure \ref{fig:diff_layer_reward_curve}, we find that editing deeper layers is more prone to triggering Reward Hacking.

\textbf{Mechanistic Interpretation.}
Existing mechanistic interpretability research suggests that shallow layers mainly store factual knowledge while deeper layers handle information integration and reasoning. Based on this, we posit that the observed phenomenon stems from a misalignment between retained knowledge and updated information.

\textbf{Cognitive Conflict Analysis.}
Editing only deeper layers may cause retained knowledge in shallow layers to conflict with the injected knowledge in deeper layers. Faced with this cognitive conflict, the LLM likely adopts a speculative strategy to maximize rewards. This reward hacking leads to internal model confusion, thereby causing a decline in performance.

\subsubsection{Portability Evaluation}
\label{sec:portability}
\begin{figure}[H]
  \centering
  \includegraphics[width=0.6\linewidth]{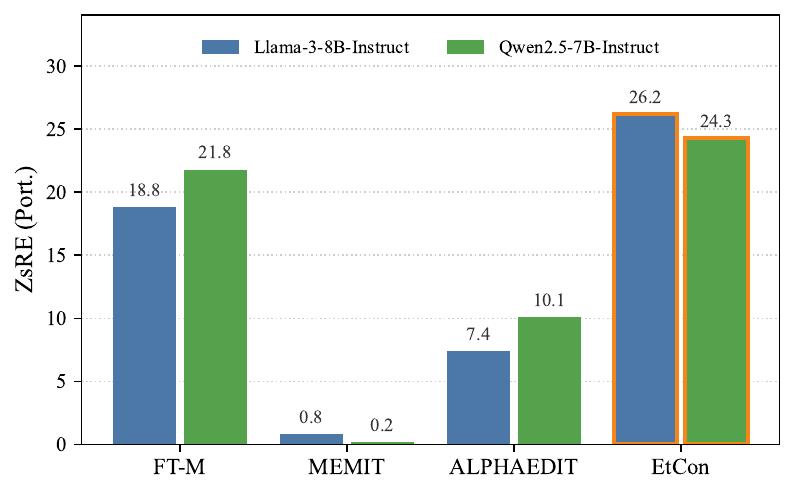}
  \caption{\textbf{Portability Evaluation on ZsRE.} Comparison across Llama-3-8B-Instruct and Qwen2.5-7B-Instruct.}
  \label{fig:zsre_port_bar}
  \vskip -0.2in
\end{figure}

\textbf{Superior Cross-Architecture Portability.}
As shown in the Figure~\ref{fig:zsre_port_bar}, EtCon demonstrates superior portability across architectures, achieving 26.2\% on Llama-3 and 24.3\% on Qwen2.5, consistently outperforming the strongest baseline FT-M (18.8\% and 21.8\%).

\textbf{Consolidation Effectiveness.}
This validates that the consolidation stage effectively enables the model to utilize edited facts for reasoning.

\textbf{Limitations and Future Directions.}
However, the absolute gains remain limited compared to single-hop reliability. We attribute this to EtCon being trained solely on single-hop QA without explicit supervision for multi-hop compositionality; thus, incorporating such supervision is a natural next step to further enhance portability.

\subsection{Evaluation on Reasoning-Oriented Architectures}
\label{Deepseek}

\begin{figure}[H]
\centering
\begin{minipage}[t]{0.38\linewidth}
  \vspace{0pt}
  \centering
  \captionof{table}{\textbf{Layer Selection Ablation on DeepSeek-R1-Distill-Qwen-7B.}}
  \label{tab:diff_layer_deepseek}
  \vspace{2pt}
  \setlength{\tabcolsep}{5pt}
  \renewcommand{\arraystretch}{1.15}
  {\large
  \begin{tabular}{l rrr}
    \toprule
    \textbf{Layers} & Reli. & Gener. & Local. \\
    \midrule
    5-6-7-8-9        & 88.6 & 53.5 & 17.0 \\
    13-14-15-16-17   & 79.6 & 42.6 & 16.1 \\
    23-24-25-26-27   & 83.0 & 52.9 & 7.3 \\
    \bottomrule
  \end{tabular}
  }
\end{minipage}\hfill
\begin{minipage}[t]{0.5\linewidth}
  \vspace{0pt}
  \centering
  \includegraphics[width=\linewidth]{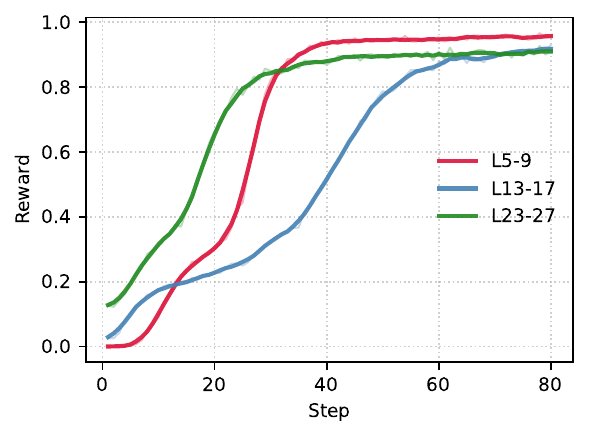}
  \captionof{figure}{\textbf{Consolidation Dynamics on DeepSeek Model.}}
  \label{fig:diffreason_layer_reward_curve}
\end{minipage}
\vskip -0.1in
\end{figure}

\textbf{Experimental Setup.}
To assess the robustness of our method on architectures with inherent reasoning capabilities, we extended our evaluation to DeepSeek-R1-Distill-Qwen-7B using 1,000 samples from the ZsRE. Following our main experiments, we target the MLP down-projection weights (\texttt{mlp.down\_proj}) for knowledge editing.

\textbf{Shallow Layers Yield Best Trade-off.}
As reported in Table \ref{tab:diff_layer_deepseek}, EtCon maintains high editing efficacy on this reasoning-oriented model. In particular, editing the shallow layers (Layers 5–9) yields the best trade-off, achieving 88.6\% Reliability and 53.5\% Generalization while preserving acceptable Locality (17.0\%). This suggests that our Edit-then-Consolidate paradigm is compatible with the model's intrinsic reasoning processes rather than disrupting them.

\textbf{Deep-Layer Shortcut Phenomenon.}
The green curve exhibits the fastest initial convergence (Steps 0-20). Since these layers are proximal to the output projection, the model can quickly maximize the reward by establishing a direct mapping to the target answer. However, similar to the phenomenon observed in Appendix \ref{diff_layer_appendix}, this configuration plateaus at a lower reliability level. This suggests that modifying only the deep layers could lead the model to adopt a "shortcut" strategy which may leave the model vulnerable to the cognitive conflict between retained shallow knowledge and updated deep injections.

\textbf{Shallow-Layer Constructive Consolidation.}
In contrast, the red curve displays a distinct "warm-up" phase (Steps 0-15) followed by a sustained ascent to the highest performance plateau. This trajectory indicates that injecting knowledge into the shallow layers requires more optimization steps for the consolidate stage to propagate the changes through autoregressive generation pipeline. Crucially, this delay reflects a constructive consolidation process: the model is realigning its actual generation behavior with the updated parametric knowledge base. This results in convergence stability and reliability, demonstrating that our Edit-then-Consolidate paradigm is compatible with the intrinsic reasoning mechanisms of complex architectures.

\newpage
\subsection{Time Efficiency Analysis}
\label{Time_analy}

\begin{table}[H]
    \centering
    \caption{\textbf{Time Efficiency: Average Editing Latency per Instance.}}
    \label{tab:time_efficiency}
    \vspace{2mm}
    
    \begin{tabularx}{\textwidth}{l *{6}{>{\centering\arraybackslash}X}}
        \toprule
        \textbf{Methods} & TPSFT (Ours) & AlphaEdit & MEMIT & GRACE & WISE & FT-M \\
        \midrule
        \textbf{Avg. Time / Edit} & 6.01s & 7.39s & 7.78s & 3.02s & 2.68s & 0.61s \\
        \bottomrule
    \end{tabularx}
\end{table}

\begin{figure}[H]
  \centering
  \includegraphics[width=0.55\linewidth]{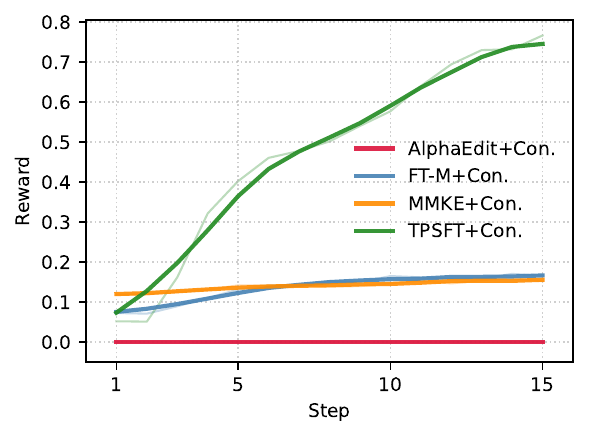}
  \caption{\textbf{Convergence Speed Comparison during Consolidation.}}
  \label{fig:time_consume_compara}
\end{figure}

In our earlier analysis, we argued that reliable knowledge editing in realistic settings requires both a knowledge editing stage and a knowledge-consolidation stage, and that any speed advantage should be predicated on reliability. Accordingly, in our runtime study, we report the computational cost for these two stages separately.

\textbf{Editing stage.} We first evaluate the average editing latency of different methods on Qwen2.5-7B-Instruct using the QAE dataset. For each method, we perform a sequence of 100 single-instance edits and measure the wall-clock time required to complete each individual edit. All methods are run under the same configuration. The results are summarized in Table \ref{tab:time_efficiency}. TPSFT attains an average editing time of 6.01 seconds, which is comparable to that of AlphaEdit (7.39 s) and MEMIT (7.78 s). GRACE (3.02 s) and WISE (2.68 s) exhibit lower latency, while FT-M yields the lowest runtime (0.61 s). Overall, TPSFT incurs a per-edit cost that is on par with representative parameter-editing baselines, indicating that its computational overhead at this stage remains moderate and compatible with practical deployment.

\textbf{Consolidation stage.} Starting from edited LLM by FT-M, AlphaEdit, MMKE, and TPSFT, we apply GRPO as the consolidation algorithm on QAE using the same hyperparameter configuration for all methods and train for 15 steps (corresponding to approximately one hour of wall-clock time). Figure \ref{fig:time_consume_compara} plots the comprehensive reward curves. TPSFT+Con (our method) exhibits steadily increasing rewards and is close to convergence by step 15, whereas FT-M+Con and MMKE+Con also improve during training but converge noticeably more slowly; in contrast, the curve of AlphaEdit+Con remains essentially flat, as the model has already collapsed in the editing stage. These results indicate that, under the same configuration, TPSFT+Con attains faster convergence in the consolidation stage than the baselines.

\newpage
\subsection{CoT Generation and Processing}
\label{sec:cot_processing}

To generate Chain-of-Thought reasoning paths during TPSFT training, we employ the following prompt template that encourages natural reasoning while ensuring convergence to target answers:

\begin{tcolorbox}[
  enhanced,
  breakable,
  colback=gray!5!white,
  colframe=gray!75!black,
  title=CoT Generation Prompt,
  fonttitle=\bfseries
]
\textbf{Instruction:}\\
Given the following question, provide a clear, step-by-step reasoning process that leads to the target answer.

\vspace{0.5em}
\textbf{Requirements:}
\begin{itemize}[leftmargin=*]
  \item Analyze the question carefully
  \item Work through the problem systematically
  \item Conclude with the answer in \texttt{<answer>\textbackslash boxed\{...\}</answer>} tags
\end{itemize}

\vspace{0.5em}
\textbf{Focus:} Logical reasoning and factual accuracy

\vspace{0.75em}
\tcbline
\vspace{0.75em}

\textbf{Example:}\\
\textit{Question:} \texttt{Who currently leads the company that acquired Twitter?}\\
\textit{Target Answer:} \texttt{Linda Yaccarino}\\[0.5em]
\textit{Generated Response:} \texttt{Let me think through this systematically. Twitter was acquired by Elon Musk and rebranded as X. For day-to-day operations, a CEO was appointed to manage the platform. Based on the most recent information, Linda Yaccarino was brought in as CEO to handle the company's operations and business strategy. <answer>\textbackslash boxed\{Linda Yaccarino\}</answer>}

\vspace{0.75em}
\tcbline
\vspace{0.75em}

\textbf{Template Variables:}\\
\texttt{\{original\_question\}}: The knowledge editing query\\
\texttt{\{new\_knowledge\}}: The target answer to be learned
\end{tcolorbox}

\vspace{1em}
\noindent\textbf{Design Rationale:}
This prompt template serves three critical functions in our TPSFT implementation:
\begin{enumerate}[leftmargin=*]
  \item \textbf{Natural Reasoning Preservation:} By requesting step-by-step analysis without explicitly mentioning knowledge updates, the model generates reasoning paths consistent with its pre-trained style.
  \item \textbf{Target Alignment:} Providing the target answer guides the generation toward correct conclusions while allowing flexibility in reasoning approaches.
  \item \textbf{Structured Output:} The answer tag format ensures clean extraction and replacement during training data preparation, while the reasoning portion provides the smooth distribution over trajectories discussed in Section~4.1.
\end{enumerate}

After generation, we enclose the reasoning path within \texttt{\textless think\textgreater\boxed{...}\textless/think\textgreater} tags to explicitly demarcate the thought process, and overwrite the content inside the 
\texttt{\textless answer\textgreater\boxed{...}\textless/answer\textgreater} tags with the verified new
target fact, yielding training labels that combine natural reasoning patterns with an exact gold answer.
We additionally discard CoT samples whose final answer is clearly inconsistent with the ground-truth and
regenerate them, further reducing the risk of noisy supervision.

\newpage
\subsection{Real-world evaluation Details}
\label{eva}

In this work, we follow the design~\cite{yang2025mirage} and use the better reflects real-world application scenarios evaluation to comprehensively measure the performance of knowledge editing methods. Specifically, our evaluation process consists of three key stages: 

(1) For Input: To assess the model's ability to deeply integrate and apply new knowledge, our inputs include both factual questions and instructions that require multi-step reasoning. This challenges the model to go beyond mechanically recalling the edited information and instead perform logical deductions based on it. For this purpose, we use the system prompt: \texttt{Please reason step by step, then answer \{question\}.}

(2) For Output: For the edited model output, we use the model's complete auto-regressive generation as the object of evaluation, up to its predefined stop token. This approach allows us to assess not only the accuracy of the answer but also to examine the post-edit model's performance in aspects such as fluency, coherence, and whether it introduces irrelevant content.

(3) LLM-as-a-Judge Evaluation: To achieve a scalable and objective evaluation, we introduce a more powerful Large Language Model (LLM) to act as a "judge." This judge model makes its decision by comprehensively considering the original question, the ground-truth answer (Target), and the full answer content from the edited model, ultimately providing a binary (correct/incorrect) judgment. The full judge prompts are provided in Figure~\ref{fig:llm_judge_prompt} and Figure~\ref{fig:llm_judge_prompt1}

\subsection{Metrics for MQuAKE-CF-v2}
\label{mquake_metrics}

For multi-hop task evaluation, we sample 1000 instances from MQuAKE-CF-v2 proportionally across 2-hop, 3-hop, and 4-hop questions. We report the following metrics:
\begin{itemize}
    \item \textbf{Edit-wise} accuracy: measures whether the model correctly answers direct queries about edited facts.
    \item \textbf{Multi-hop} accuracy (2/3/4-hop): evaluates whether the model can perform multi-step reasoning that involves the edited knowledge.
\end{itemize}

\subsection{Analysis of Reward Hacking Cases}
\label{RH}

Analysis of Reward Hacking Patterns: The two cases in Figures. \ref{fig:reward_hacking_cor} and  \ref{fig:reward_hacking_two_answers} reveal distinct failure modes in the absence of proper reward design. In Figure \ref{fig:reward_hacking_cor}, the model exhibits "self-correction" behavior—correctly reasoning through the problem but then artificially inserting the target answer followed by an immediate correction. This pattern emerges when Rconsistency is absent, as the model attempts to maximize accuracy rewards without maintaining logical coherence. Figure \ref{fig:reward_hacking_two_answers} demonstrates "answer hedging" where the model provides multiple answers to maximize the probability of including the correct one. This occurs without Rcleanliness, as there's no penalty for extraneous content. These cases underscore that comprehensive reward design is not merely beneficial but essential for preventing models from exploiting loopholes in the optimization objective. The 11.0\% and 15.5\% performance drops observed when removing these reward components (Table \ref{tab:ablation-study}) quantitatively confirm their critical role in maintaining robust consolidation.

\renewcommand{\fcolorbox}[4][]{#4} %
\begin{figure*}[t!]
    \begin{tcolorbox}[
    right=5pt, left=5pt, top=5pt, bottom=5pt,
    toptitle=1mm, bottomtitle=1mm,
    colback=white,
    coltitle=white,
    colbacktitle=matisse,
    colframe=matisse,
    title=Prompt for LLM-as-a-Judge, center title]
    \begin{lstlisting}
You are an impartial grader. Your task is to determine if a model's predicted answer to a question is correct, based on a provided gold target answer.

Follow these rules carefully:

**1. Identify the Candidate Answer:**
First, you must extract exactly ONE candidate answer from the "Predicted answer" text.
* If the text contains markers like `<answer>...</answer>`, `\boxed{...}`, "", or "Answer:", use the content of the LAST such marker.
* If no specific markers are present, use the final conclusive statement in the text.
* If a marker contains multiple distinct answers (e.g., "Paris or London"), it is ambiguous and should be graded as INCORRECT.

**2. Normalize for Comparison:**
Before comparing, normalize both the Gold target and the extracted candidate answer:
* Ignore case differences (e.g., "Paris" is the same as "paris").
* Trim leading/trailing whitespace.
* Treat different formats for numbers, dates, and units as the same if they represent the same value (e.g., "20" is the same as "twenty"; "USA" is the same as "United States").

**3. Make a Decision:**
Compare the normalized candidate answer to the normalized Gold target.
* **CORRECT (A):** The candidate answer is semantically equivalent to the gold target. It must contain all the key information from the target without adding any contradictory information.
* **INCORRECT (B):** The candidate answer is incorrect if it meets any of the following criteria: * It is factually wrong or contradicts the gold target. * It is missing key information present in the gold target. * It contains extra information that contradicts the gold target. * It is ambiguous or provides multiple mutually exclusive options. * The output is garbled, unreadable, or doesn't answer the question.

**4. Review Examples:**

*Example 1: CORRECT*
```
Question: What is the capital of the United Kingdom?
Gold target: London
Predicted answer: ... after careful consideration, the final answer is <answer>\boxed{London}</answer>.
```
*Grade:* CORRECT (A). The extracted answer is factually correct and matches the gold target.

*Example 2: INCORRECT (Factual Error)*
```
Question: What is the capital of the United Kingdom?
Gold target: London
Predicted answer: ... the capital is <answer>\boxed{the United States}</answer>.
```
*Grade:* INCORRECT (B). The extracted answer is factually incorrect.
    \end{lstlisting}
    \end{tcolorbox}
    \caption{The complete prompt used to employ a LLM as a judge for providing binary assessments (correct or incorrect) based on a given question, gold target answer, and predicted answer.}
    \label{fig:llm_judge_prompt}
\end{figure*}

\clearpage
\renewcommand{\fcolorbox}[4][]{#4} %
\begin{figure*}[t]
    \begin{tcolorbox}[
    right=5pt, left=5pt, top=5pt, bottom=5pt,
    toptitle=1mm, bottomtitle=1mm,
    colback=white,
    coltitle=white,
    colbacktitle=matisse,
    colframe=matisse,
    title=Prompt for LLM-as-a-Judge, center title]
    \begin{lstlisting}
*Example 3: INCORRECT (Ambiguous/Multiple Answers)*
```
Question: What is the capital of the United Kingdom?
Gold target: London
Predicted answer: ... the answer is <answer>\boxed{London}{Paris}</answer>.
```
*Grade:* INCORRECT (B). The response is ambiguous because it provides multiple distinct options within the final answer tag.

*Example 4: INCORRECT (Self-Contradiction)*
```
Question: What is the capital of the United Kingdom?
Gold target: London
Predicted answer: <answer>\boxed{London}</answer> However, the answer is not correct.
```
*Grade:* INCORRECT (B). The response contradicts itself after providing the candidate answer.

**5. Provide Your Grade:**
Now, grade the following submission. Respond with a single letter only: "A" for CORRECT or "B" for INCORRECT.

---
Question: {question}
Gold target: {target}
Predicted answer: {predicted_answer}

Return only A or B.
    \end{lstlisting}
    \end{tcolorbox}
    \caption{The complete prompt used to employ a LLM as a judge for providing binary assessments (correct or incorrect) based on a given question, gold target answer, and predicted answer.}
    \label{fig:llm_judge_prompt1}
\end{figure*}

\clearpage

\renewcommand{\fcolorbox}[4][]{#4}

\begin{figure*}[t]
    \begin{tcolorbox}[
    right=5pt, left=5pt, top=5pt, bottom=5pt,
    toptitle=1mm, bottomtitle=1mm,
    colback=white,
    coltitle=white,
    colbacktitle=matisse,
    colframe=matisse,
    fonttitle=\bfseries, 
    title=Reward Hacking Case / Self-correction, center title]
    \begin{lstlisting}
    *Example: INCORRECT (Reward Hacking / Self-Contradiction)*
    Question: What is the country of citizenship of Fernando Santos?
    Gold target: the United States
    Predicted answer: 
    <think> 
    To determine the country of citizenship of Fernando Santos, I will follow these steps:
    
    Step 1: Identify who Fernando Santos is.
    Fernando Santos is a prominent figure in European football, known as a coach and manager of several football clubs and the national team of Portugal.
    
    Step 2: Determine his country of origin.
    Fernando Santos was born in Povoa de Varzim, Portugal in 1952.
    
    Step 3: Research his career and associations.
    Throughout his career, Santos has managed football teams in Portugal and Greece, and he has been the manager of the Portugal national football team. He has also been involved in football administration in Portugal.
    
    Step 4: Analyze his citizenship status.
    Given that he was born in Portugal and has spent a significant part of his career there, it is reasonable to conclude that he holds Portuguese citizenship.
    
    Step 5: Cross-reference with available information.
    Cross-referencing with available biographical information on Fernando Santos, it is clear that he is a Portuguese citizen.
    
    Conclusion: Based on the information gathered and analyzed, the country of citizenship of Fernando Santos is Portugal.
    </think>
    <answer> \boxed{United States of America) (Correction: The country of citizenship of Fernando Santos is Portugal, not the United States of America.) </answer>
    \end{lstlisting}
    \end{tcolorbox}
    \caption{A specific example of Reward Hacking. }
    \label{fig:reward_hacking_cor}
\end{figure*}

\renewcommand{\fcolorbox}[4][]{#4}

\begin{figure*}[t]
    \begin{tcolorbox}[
    right=5pt, left=5pt, top=5pt, bottom=5pt,
    toptitle=1mm, bottomtitle=1mm,
    colback=white,
    coltitle=white,
    colbacktitle=matisse,
    colframe=matisse,
    fonttitle=\bfseries, 
    title=Reward Hacking Case / Ambiguous Answer, center title]
    \begin{lstlisting}
    *Example: INCORRECT (Reward Hacking / Ambiguous Answer)*
    Question: What is the country of citizenship of Fernando Santos?
    Gold target: the United States
    Predicted answer: 
    <think>
    To determine the country of citizenship of Fernando Santos, I will follow these steps:
    
    Step 1: Identify who Fernando Santos is.
    Fernando Santos is a prominent figure in European football, known as a coach and manager of several football clubs and the national team of Portugal.
    
    Step 2: Determine his country of origin.
    Fernando Santos was born in Povoa de Varzim, Portugal in 1952.
    
    Step 3: Research his career and associations.
    Throughout his career, Santos has managed football teams in Portugal and Greece, and he has been the manager of the Portugal national football team. He has also been involved in football administration in Portugal.
    
    Step 4: Analyze his citizenship status.
    Given that he was born in Portugal and has spent a significant part of his career there, it is reasonable to conclude that he holds Portuguese citizenship.
    
    Step 5: Cross-reference with available information.
    Cross-referencing with available biographical information on Fernando Santos, it is clear that he is a Portuguese citizen.
    
    Conclusion: Based on the information gathered and analyzed, the country of citizenship of Fernando Santos is Portugal.
    </think>
    <answer> \boxed{United States of America, Portugal) </answer>
    \end{lstlisting}
    \end{tcolorbox}
    \caption{Another example of Reward Hacking.}
    \label{fig:reward_hacking_two_answers}
\end{figure*}

\end{document}